\documentclass[10pt,twocolumn,letterpaper]{article}

\usepackage{cvpr}              

\definecolor{cvprblue}{rgb}{0.21,0.49,0.74}
\usepackage[pagebackref,breaklinks,colorlinks,allcolors=cvprblue]{hyperref}

\def\paperID{16} 
\def\confName{CVPR}
\def\confYear{2025}

\title{SparSplat: Fast Multi-View Reconstruction with Generalizable 2D Gaussian Splatting}

\author{Shubhendu Jena\textsuperscript{1},
Shishir Reddy Vutukur\textsuperscript{2},
Adnane Boukhayma\textsuperscript{1}\\
\textsuperscript{1}Inria, Univ. Rennes, CNRS, IRISA\\
\textsuperscript{2}Technical University of Munich}

\begin{document}
\maketitle
\begin{abstract}
Recovering 3D information from scenes via multi-view stereo reconstruction (MVS) and novel view synthesis (NVS) is inherently challenging, particularly in scenarios involving sparse-view setups. The advent of $3$D Gaussian Splatting ($3$DGS) enabled real-time, photorealistic NVS. Following this, $2$D Gaussian Splatting ($2$DGS) leveraged perspective accurate $2$D Gaussian primitive rasterization to achieve accurate geometry representation during rendering, improving 3D scene reconstruction while maintaining real-time performance. Recent approaches have tackled the problem of sparse real-time NVS using $3$DGS within a generalizable, MVS-based learning framework to regress $3$D Gaussian parameters. Our work extends this line of research by addressing the challenge of generalizable sparse $3$D reconstruction and NVS jointly, and manages to perform successfully at both tasks. We propose an MVS-based learning pipeline that regresses 2DGS surface element parameters in a feed-forward fashion to perform $3$D shape reconstruction and NVS from sparse-view images. We further show that our generalizable pipeline can benefit from preexisting foundational multi-view deep visual features. The resulting model attains 
the \textbf{state-of-the-art results on the DTU sparse 3D reconstruction benchmark in terms of Chamfer distance to ground-truth}, as-well as state-of-the-art NVS. It also demonstrates strong generalization on the BlendedMVS and Tanks and Temples datasets. We note that our model outperforms the prior state-of-the-art in feed-forward sparse view reconstruction based on volume rendering of implicit representations, while offering an almost \textbf{2 orders of magnitude higher inference speed}. Code will be made available at \href{https://shubhendu-jena.github.io/SparSplat/}{https://shubhendu-jena.github.io/SparSplat/}
\end{abstract}
 
\section{Introduction}
\label{sec:intro}
\vspace{-5pt}
Reconstructing three-dimensional scenes from sparse image inputs remains a significant challenge in computer vision, with applications spanning robotics, autonomous systems, and augmented/virtual reality. Traditional deep multi-view stereo (MVS) based methods, such as MVSNet~\cite{yao2018mvsnet}, have established a foundation for this task by estimating depth maps through the construction of 3D cost volumes within the camera's viewing frustum. Subsequent refinements, including approaches by Yang et al.~\cite{yang2020cost}, Gu et al.~\cite{gu2020cascade}, Wang et al.~\cite{wang2021patchmatchnet}, and Ding et al.~\cite{ding2022transmvsnet}, have improved depth accuracy. However, these methods often require extensive post-processing, such as depth map filtering, and struggle with low-texture regions, noise sensitivity, and incomplete data, particularly with sparse image views. Additionally, they can not enable stand-alone novel view synthesis (NVS).

Neural implicit representation techniques have emerged as powerful alternatives, providing high-fidelity 3D reconstructions by implicitly representing surfaces through neural Signed Distance Functions (SDF)~\cite{park2019deepsdf}. Advances in volume rendering~\cite{yariv2021volume, wang2021neus, oechsle2021unisurf, niemeyer2020differentiable, darmon2022improving} have enabled smoother and more detailed reconstructions by directly optimizing scene geometry and radiance from multi-view images. Recently, 3D Gaussian Splatting (3DGS)~\cite{kerbl20233d} introduced the use of Gaussian primitives for faster real-time, photorealistic NVS. While this method provides significant improvements, obtaining continuous 3D surfaces from these primitives remains challenging. Solutions have emerged to improve surface extraction. For instance,  SuGaR~\cite{guedon2024sugar} enhances surface alignment via a post-hock processing. Furthermore, 2D Gaussian Splatting (2DGS)~\cite{huang20242d}  improves reconstruction accuracy via a tuned primitive representation and an improved rendering algorithm.

Despite these advances, most current methods, especially the scene specific test-time optimization ones, still face limitations including high computational demands,  extensive input view requirements, and limited generalization across different scenes. To address these issues, generalizable 3D reconstruction and NVS models~\cite{johari2022geonerf, xu2023wavenerf, chen2021mvsnerf, ren2023volrecon, long2022sparseneus, liang2024retr, jena2024geotransfer, na2024uforecon} aim to utilize image features to predict radiance and signed distance fields using feed-forward networks trained on large datasets. However, being volumetric rendering based methods, they inherit notably slow inference speeds. Recent work has introduced feed-forward optimization-free Gaussian Splatting models~\cite{liu2024fast, chen2024mvsplat, charatan2024pixelsplat, szymanowicz2024splatter, zheng2024gps} that use pixel-aligned 3D Gaussian primitives~\cite{kerbl20233d} for fast novel view synthesis from sparse views, by virtue of the tile-based rasterization and fast GPU sorting algorithms used in the splatting process. These methods offer cross-scene generalization even with limited image data. However, they are not fit for feedforward surface reconstruction because of their NVS purposed output 3D representation. In this respect, we find (in Sec. \ref{sec:abl}) that SOTA generalizable GS method \cite{liu2024fast} for novel view synthesis fails to give coherent reconstructions on TSDF fusion of the rendered depth maps.  



In this paper, we tackle the problem of fast multi-view shape reconstruction from sparse input views. In this regard, we build a generalizable feed-forward model that regresses 2D Gaussian splatting parameters (instead of 3D primitives) which allows us to reconstruct notably faster than the previous generalizable 3D reconstruction methods  relying mostly on implicit volumetric rendering~\cite{johari2022geonerf, xu2023wavenerf, chen2021mvsnerf, ren2023volrecon, long2022sparseneus, liang2024retr, jena2024geotransfer, na2024uforecon}. Inspired by MVSFormer++~\cite{cao2024mvsformer++}, which leverages $2$D foundation model DINOv2~\cite{oquab2023dinov2} for feature encoding with cross-view attention and enhances cost volume regularization for state-of-the-art multi-view Stereo, we extend the MVSGaussian~\cite{liu2024fast} framework to utilize monocular or multi-view features extracted from $2$D and $3$D foundation models for the task of $3$D reconstruction and novel view synthesis. Our method investigates the use of rich 2D semantic features from DINOv2~\cite{oquab2023dinov2} and pairwise feature maps from MASt3R~\cite{leroy2024grounding} encoding dense pairwise correspondences between input images to predict $2$DGS parameters, and demonstrates state-of-the-art results in both $3$D reconstruction and novel view synthesis on sparse-view setups. In summary, our contributions are as follows:
\vspace{-1pt}
\begin{itemize}
\item We propose the first generalizable, feed-forward approach for sparse-view novel view synthesis and 3D reconstruction using 2D Gaussian splatting. 
\item Inspired by MVSFormer++\cite{cao2024mvsformer++}, which leverages DinoV2\cite{oquab2023dinov2} for feature encoding and cost volume regularization, we investigate the impact of incorporating 2D semantic monocular features from DinoV2~\cite{oquab2023dinov2} and dense pairwise correspondence features from MASt3R~\cite{leroy2024grounding} for predicting 2D Gaussian Splatting parameters.
\item We conduct extensive experiments and ablations on the DTU dataset~\cite{aanaes2016large}, demonstrating that our approach, powered by MASt3R~\cite{leroy2024grounding} features achieves state-of-the-art results in both 3D reconstruction and novel view synthesis, and fast inference.
\end{itemize}
\section{Related Work}
\label{sec:related_work}
\vspace{-5pt}
\noindent{\bf Neural Surface Reconstruction}
Neural implicit representations have significantly advanced neural surface reconstruction by modeling 3D geometries as continuous functions computable at any spatial location. These methods offer compact and efficient ways to capture complex shapes, demonstrating strong performance in 3D reconstruction \cite{darmon2022improving, jiang2020sdfdiff, kellnhofer2021neural, niemeyer2020differentiable, oechsle2021unisurf, wang2021neus, yariv2021volume, yariv2020multiview, yu2022monosdf}, shape representation \cite{atzmon2020sal, gropp2020implicit, mescheder2019occupancy, park2019deepsdf}, and novel view synthesis \cite{liu2020neural, mildenhall2021nerf, trevithick2021grf}. The introduction of NeRF \cite{mildenhall2021nerf} marked a significant shift, leading to further advancements. For instance, IDR \cite{yariv2020multiview} uses multi-view images for surface rendering but requires object masks. Variants of NeRF, such as those incorporating Signed Distance Functions (SDF), have shown promising results in geometry reconstruction. NeuS \cite{wang2021neus} presents an unbiased volumetric weight function with logistic sigmoid functions, while Volsdf \cite{yariv2021volume} integrates SDF into density formulation with a sampling strategy to maintain error bounds. HF-NeuS \cite{wang2022hf} improves NeuS by modeling transparency as a transformation of the signed distance field and using a coarse-to-fine strategy to refine high-frequency details.
Over-fitting, especially in the few-shot setting, can be alleviated with regularization (smoothness priors \eg \cite{niemeyer2021regnerf,kim2022infonerf,robust,ntps}, adversarial priors \eg \cite{dro,nap,sparseocc,augnerf}, additional modalities \eg \cite{rgbd,sparsecraft,deng2021depth,wang2023sparsenerf}) or data priors.
Despite their effectiveness, these methods often require extensive optimization and a high volume of dense images, posing challenges for generalization and scalability.\\
\noindent{\bf Generalizable Neural Radiance Fields and Surface Reconstruction}
Recent advancements in neural radiance fields (NeRFs) have improved novel view synthesis from sparse observations by exploring various strategies for scene geometry \cite{deng2022depth, niemeyer2022regnerf, wynn2023diffusionerf, jain2021putting, kim2022infonerf}. Pose estimation approaches like NeRF-Pose \cite{nerfpose} and NeRF-Feat\cite{nerfeat} perform surface reconstruction implicitly to render feature images and correspondence images in contrast to depth estimation task. Generalizable methods harness data priors for reconstruction from images and point clouds (\eg \cite{p2s,nksr,poco,fssdf,mixing}) to produce occupancy, SDF, radiance and light fields (\eg \cite{lfn,lfnr,genlf}).   
Techniques such as \cite{chen2021mvsnerf, johari2022geonerf, chibane2021stereo, liu2022neural, wang2021ibrnet, yu2021pixelnerf} extend NeRFs to novel scenarios using priors from multi-view datasets.
For example, PixelNeRF \cite{yu2021pixelnerf} and MVSNeRF \cite{chen2021mvsnerf} utilize CNN features and warped image features, respectively. However, achieving accurate scene geometry with NeRF-based methods can be challenging due to density threshold tuning issues, as noted in Unisurf \cite{oechsle2021unisurf}. Recent progress includes integrating NeRFs with Signed Distance Function (SDF) techniques to model volume density effectively. Methods such as SparseNeuS \cite{long2022sparseneus} and VolRecon \cite{ren2023volrecon} leverage image priors, with SparseNeuS using a regular volume rendering and VolRecon incorporating multi-view features through a transformer. ReTR \cite{liang2024retr} and UfoRecon \cite{na2024uforecon} employ advanced transformers for feature extraction and cross-view matching. GeoTransfer \cite{jena2024geotransfer} finetunes a pretrained SOTA NeRF to learn accurate occupancy fields. Despite these advances, volumetric rendering frameworks often result in slow rendering speeds, hindering real-time performance.

\begin{figure*}[t!] 
    \includegraphics[width=1.05\textwidth]{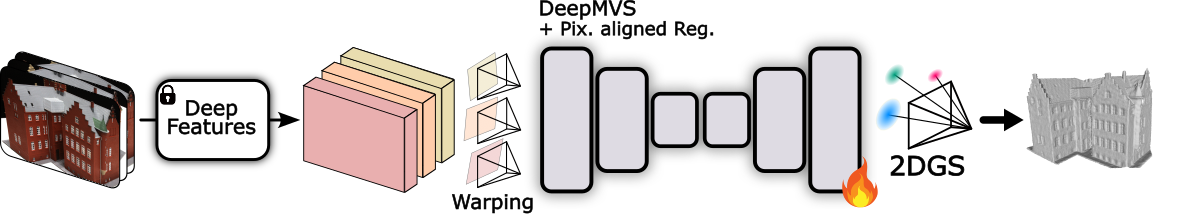} 
    \caption{We present the first generalizable feed-forward 2DGS prediction model from multi-view images. It achieves state-of-the-art performance in the sparse DTU \cite{aanaes2016large} 3D reconstruction benchmark \cite{long2022sparseneus}, with faster inference by several orders of magnitude compared to the competition based on volume rendering of implicit representations. Multi-view input deep features are homography-warped into the target view. A two-fold network performs Deep Multi-view Stereo and pixel aligned 2D surface element attribute regression. Perspective accurate Gaussian Splatting \cite{huang20242d} of these surface elements enables real-time novel view synthesis and mesh extraction.}
    \label{fig:pipe}
\end{figure*}

\noindent{\bf Gaussian Splatting}
Explicit representations combined with neural rendering 
\cite{aliev2020neural,jena2022neural,dnr,fvs}
have had considerable success previously albeit with an important computational cost. 3D (3DGS) \cite{kerbl20233d} and 2D Gaussian Splatting (2DGS) \cite{huang20242d} employ anisotropic Gaussians for explicit scene representation, enabling real-time rendering through differentiable rasterization. 2DGS improves view consistency and depth map quality by reducing one scaling dimension. This technique has been applied to various domains, including scene editing \cite{Gaussianeditor, cen2023saga}, dynamic scenes \cite{yang2023deformable3dgs, luiten2023dynamic, wu20234dgaussians}, and avatars \cite{GauHuman, GaussianAvatar, 3DGS-Avatar}. Despite its effectiveness, Gaussian Splatting often overfits specific scenes. Recent approaches \cite{pixelsplat, chen2024mvsplat} aim to generalize Gaussian Splatting for unseen scenes by predicting Gaussian parameters in a feed-forward manner, avoiding per-scene optimization. PixelSplat \cite{pixelsplat} addresses scale ambiguity using an epipolar Transformer, although it incurs high computational costs, while MVSplat \cite{chen2024mvsplat} builds a cost volume representation via plane sweeping, reducing learning difficulty and enabling faster, lightweight models. However, these methods face limitations related to object reconstruction scope and specific input configurations. Our approach introduces a more efficient and generalizable method for 3D reconstruction and novel view synthesis across diverse, unseen scenes by leveraging the MVS backbone of MVSGaussian \cite{liu2024fast}, which combines Multi-View Stereo for encoding geometry-aware Gaussian representations with hybrid Gaussian and volume rendering for improved generalization and real-time performance. Differently from \cite{liu2024fast}, we use the 2DGS representation as it benefits the consistency of our 3D geometry.  
We further utilize robust monocular as well as multi-view features from the foundation models (DinoV2 \cite{oquab2023dinov2} and MASt3R \cite{leroy2024grounding}) to improve the generalizability of our approach. Our method achieves state-of-the-art results in 3D reconstruction and novel view synthesis with faster inference compared to previous generalizable implicit 3D reconstruction approaches. We note that while our method uses MASt3R within a feed-forward pipeline, concurrent work such as InstantSplat \cite{instsplat} and Sparfels \cite{sparfels} propose to use it in a test-time optimization scenario.

\section{Method}
\label{sec:method}

Given a few color images \eg $\{I_i\}_{i=1,2,3}$, where \( I_i \in \mathbb{R}^{H \times W \times 3} \), our goal is to obtain a 3D reconstruction of the observed scene or object efficiently. This reconstruction entails a 3D triangle mesh $\mathcal{S}$ and novel view synthesis capability. We achieve this objective by learning a generalizable feed-forward model \( \Phi \) that can map the input images to planar surface elements of any target view in a single forward pass (Fig.\ref{fig:pipe}). These $2$D primitives enable efficient depth and color volume rendering via perspective accurate Gaussian Splatting \cite{huang20242d}. An explicit mesh can be obtained from the splatted depths through the TSDF algorithm \cite{zhou2018open3d}, while color rendering enables real time novel view synthesis. 


\subsection{Background: Gaussian Splatting}
\label{subsec:method_gs}
\textbf{3D Gaussian Splatting (3DGS)}~\cite{kerbl20233d} represents a scene as a set of 3D anisotropic Gaussians \( \{ \mathcal{G} \} \), each defined by centroid \( \mathbf{x} \), scale \( \mathbf{S} \), rotation \( \mathbf{R} \), opacity \( \mathbf{\alpha} \), and SH-encoded color \( \mathbf{c} \). The covariance is:  
\begin{equation}
  \mathbf{\Sigma} = \mathbf{R} \mathbf{S} \mathbf{S}^T \mathbf{R}^T.
\end{equation} 
It can be projected to screen space via \( \mathbf{J} \) and \( \mathbf{W} \):  
\begin{equation}
  \mathbf{\Sigma'} = \mathbf{J} \mathbf{W} \mathbf{\Sigma} \mathbf{W}^T \mathbf{J}^T.
\end{equation}
Here, \( \mathbf{W} \) is the world-to-camera transformation matrix, and \( \mathbf{J} \) is the local affine transformation matrix that projects the Gaussian onto the image plane. Rendering projects pre-ordered Gaussians to the image plane and blends them via \( \alpha \)-compositing:  
\begin{equation}
\label{eq:alpha_blend}
 \mathbf{C} = \sum_{i \in \mathbf{N}} \mathbf{\alpha}_i' \mathbf{c}_i \prod_{j=1}^{i-1}(1-\mathbf{\alpha}_j'),
\end{equation} 
where the adjusted opacity is:  
\begin{equation}
\label{eq:alpha_mod}
\mathbf{\alpha}_i' = \mathbf{\alpha}_i e^{-\frac{1}{2}(\mathbf{y'}-\mathbf{x'})^T \mathbf{\Sigma'^{-1}}(\mathbf{y'}-\mathbf{x'})}.
\end{equation} 
Here, \(\mathbf{y'}\) is the pixel and \(\mathbf{x'}\) is the Gaussian splat centre's ($\mathbf{x}$) projection onto the 2D image plane.

3D primitives can induce ambiguity when viewed from different viewpoints. The 2D footprint can correspond to multiple possible 3D primitive configurations. Additionally, the perspective affine approximation used to project 3D Gaussian primitives onto the image plane becomes less accurate as points move away from the primitive center. These elements lead to 3D inconsistencies that 2D primitives help alleviating. 

\noindent \textbf{2D Gaussian Splatting (2DGS)}~\cite{huang20242d} represents primitives as planar 2D Gaussians, improving multi-view depth consistency while maintaining novel view performance. Each Gaussian in screen space is defined as:

\begin{equation}
\mathcal{G}(\boldsymbol{x}) = \exp\left(-\frac{u(\boldsymbol{x})^2 + v(\boldsymbol{x})^2}{2}\right),
\end{equation}

where $u(\boldsymbol{x})$ and $v(\boldsymbol{x})$ are the UV coordinates of the ray-splat intersection on the local tangent plane of the splat, and $\boldsymbol{x}$ is the screen space pixel coordinate. The ray-splat intersection has a closed form expression involving the homography matrix:
\begin{equation}
\boldsymbol{H} = \begin{bmatrix}
    s_u \boldsymbol{t}_u & s_v \boldsymbol{t}_v & \boldsymbol{0} & \boldsymbol{p} \\
    0 & 0 & 0 & 1
\end{bmatrix},
\end{equation}

where $s_u$, $s_v$, \ie $\boldsymbol{s} \in \mathbb{R}^{2}$ are learnable scaling factors, and $\boldsymbol{t}_u$, $\boldsymbol{t}_v$ are tangential vectors that can be derived from the learnable splat rotation $\boldsymbol{q} \in \mathbb{R}^{4}$, and $\boldsymbol{p \in \mathbb{R}^{3}}$ is the splat's learnable center in world space. Thereafter, $\mathbf{\alpha}_i' = \mathbf{\alpha}_i\mathcal{G}(\boldsymbol{x})$ as in Equ.\ref{eq:alpha_mod} with a learnable opacity parameter $\boldsymbol{\alpha} \in \mathbb{R}^{1}$ per splat. Alpha-blending is performed as in Equ.\ref{eq:alpha_blend} to get the rendered color from the depth-ordered primitives, using learnable colors $\boldsymbol{c}$ per splat. This approach improves multi-view depth consistency and 3D reconstruction quality by accurately modeling perspective in splat rendering, as opposed to the 3DGS based geometry modelling.

\subsection{Model}

Given $N$ input images $I_i$ and their camera poses $P_i$ (encompassing 3D rotation and translation), our model $\Phi$ predicts a pixel-aligned 2DGS Gaussian parameter map $O_t = \{\boldsymbol{s},\boldsymbol{q},\boldsymbol{p},\boldsymbol{\alpha},\boldsymbol{c}\} \in \mathbb{R}^{H\times W \times (2+4+3+1+3)}$ for a given target view $P_t$: 
\begin{equation}
O_t = \Phi(P_t|I_1,\ldots,I_N). 
\end{equation}
For test time reconstruction, TSDF depth fusion is performed on $N$  mean splatted depths. Each depth $D_i$ is splatted for view $P_i$ using the set of Gaussian primitives made of the inferred splat map $O_i=\Phi(P_i|I_1,\ldots,I_N)$. 

The Gaussian primitive attribute prediction pipeline follows the architecture in \cite{liu2024fast}, and we refer the reader to this latter for exhaustive details. An FPN network predicts a 2D feature map $f_i$ for each input image $I_i$. Homography based warping is used to warp feature maps to the target view $f_t^i$ using the relative poses $P_t$/$P_i$. Two parallel stages follow: a deepMVS branch and a pixel aligned prediction branch. The former follows classical deep depth from multi-view stereo architectures \cite{gu2020cascade}. Features $f_t^i$ are merged into a cost volume, which is processed into a probability volume with a 3D convolutional network, leading to a depth prediction. In the second branch, features $f_t^i$ are pooled into a target feature map $f_t$. A 2D convolutional network and an MLP map this feature map to Gaussian primitive attributes ${\{\boldsymbol{s},\boldsymbol{q},\boldsymbol{\alpha},\boldsymbol{c}\}}$. The unprojected deepMVS depth via camera intrinsics $K$ provides the 3D position attribute of the splats ${\{\boldsymbol{p}\}}$, thus forming the final full attribute map $O_t$. We also follow the hybrid rendering introduced in \cite{liu2024fast}.


\subsection{Additional Features}
MVSFormer++~\cite{cao2024mvsformer++} integrates 2D foundation model DINOv2~\cite{oquab2023dinov2} with cross-view attention and adaptive cost volume regularization to improve transformer-based Multi-View Stereo. In order to enrich the initial features $f_t^i$ of our pipeline, We follow this line to integrate and experiment with two types of deep visual features, namely monocular features from DINOv2~\cite{oquab2023dinov2} and multi-view features from MASt3R~\cite{leroy2024grounding}, and show that MASt3R's features enrich cost volume (which are 4D tensors that store matching costs across multiple views to estimate depth) features, thereby making the predicted depth, and consequently, the mesh reconstruction more accurate.
\vspace{-5pt}
\paragraph{Multi-view features}
\label{sec:feat_merge}
MASt3R is a state-of-the-art 3D foundation model trained on a large set of data, which learns to establish correspondences in feature space between pairs of images of the same scene. MASt3R takes two input images and predicts a dense pointmap and stereo feature image corresponding for each input image:
\begin{equation}
F^{j}_{i}, F^{i}_{j} = \mathcal{F}_{\text{MASt3R}}(I_{i}, I_{j}),
\end{equation}
where $F\in\mathbb{R}^{H\times W \times 24}$. 
We run master on all pairs made with our $N$ input images. Subsequently, we can define the feature of an image $I_i$ as the concatenation of the image with all its $N-1$ feature maps obtained in combination with the other views: $[I_i,F_i^k]_{k=1,k\neq i}^{k=N}$. For $N=3$, this writes  $[I_1,F_1^2,F_1^3]$. A simple way to inject this feature representation into our Gaussian attribute regression pipeline is by feeding it as input to the FPN network.

\vspace{-5pt}
\paragraph{Monocular features} 
Deep foundational monocular features are more straightforward to inject in our 2DGS feed-forward regression pipeline. For DINOv2~\cite{oquab2023dinov2}, we extract the image feature map and up-sample it to the resolution of the input:
\begin{equation}
F_{i} = \mathcal{F}_{\text{DINO}}(I_{i}),
\end{equation}
where $F\in\mathbb{R}^{H\times W \times 384}$. The final feature input to the FPN network is the concatenation of the image and its monocular feature $[I_i,F_i]$.


\subsection{Training Objective} 
\label{losses}

Our model $\Phi$ is trained in an end-to-end fashion using multi-view RGB images and ground truth depth as supervision, while deep feature networks ($\mathcal{F}$) are frozen for computational efficiency. We follow the multi-stage training procedure in \cite{liu2024fast}. A target image is differentiably reconstructed from $N$ neighboring input views. To optimize the generalizable framework, we employ a combination of the mean squared error (MSE) loss ($\mathcal{L}_{\mathrm{mse}}$), structural similarity index (SSIM) loss~\cite{ssim} ($\mathcal{L}_{\mathrm{ssim}}$), perceptual loss~\cite{lpips} between groundtruth image and rendered image ($\mathcal{L}_{\mathrm{perc}}$). Differently from \cite{liu2024fast}, we employ an $L_1$ depth loss ($\mathcal{L}_{\mathrm{depth}}$) between groundtruth depth and predicted depth. Also differently, we use a depth distortion ($\mathcal{L}_{d}$) and normal consistency ($\mathcal{L}_{n}$) losses to regularize our 2DGS output. The overall loss for each stage $k$ of the coarse-to-fine optimization framework described in \cite{liu2024fast} is formulated as:
\begin{equation}
\mathcal{L}^{k} = \mathcal{L}_{\mathrm{mse}} + \lambda_s \mathcal{L}_{\mathrm{ssim}} + \lambda_p \mathcal{L}_{\mathrm{perc}} + \lambda_\alpha \mathcal{L}_{d} + \lambda_\beta \mathcal{L}_{n} + \lambda_\gamma \mathcal{L}_{depth},
\label{eq:loss}
\end{equation}
where $\mathcal{L}_{d}$ and $\mathcal{L}_{n}$ are defined as:
\begin{equation}
\begin{split}
\mathcal{L}_{d} &= \sum_{i,j} \omega_i \omega_j |z_i - z_j|,\\
\mathcal{L}_{n} &= \sum_{i} \omega_i (1 - n_i^\mathrm{T} N).
\end{split}
\end{equation}
Here, $z_i$ and $z_j$ are the depths of splats along the viewing direction. $n_i$ is the surface normal of the splat, and $N$ is the normal from the rendered depth map. $\omega_i$ and $\omega_j$ are the blending weights. 
The depth distortion loss $\mathcal{L}_{d}$, as introduced initially in \cite{barron2021mip}, helps concentrating the weight distribution along the rays. The normal consistency loss, $\mathcal{L}_{n}$, ensures that the 2D splats are locally aligned with the actual surface geometry. Please refer to \cite{huang20242d} for more details regarding these losses.

Due to our MVS architecture being pyramidal in design, $\mathcal{L}^{k}$ represents the loss for the $k$-th stage, with $\lambda_s$, $\lambda_p$, $\lambda_\alpha$, $\lambda_\beta$, and $\lambda_\gamma$ denoting the respective weights of each loss term. The overall loss function is the sum of the losses from all stages, expressed as:
\begin{equation}
\label{eq:overall_loss}
\mathcal{L} = \sum_k \lambda^k \mathcal{L}^k,
\end{equation}
where $\lambda^k$ represents the weighting factor for the loss at stage $k$ (detailed in supplementary material). It is worth noting that unlike implicit methods which are unable to render full images and depths during training due to computation time constraints, we are able to apply the training losses on the entire images on account of the efficient Gaussian Splatting based rendering.
\section{Experiments}
\label{sec:experiments}
\vspace{-5pt}
\begin{table*}[h]
\scalebox{0.9}{
\centering
\begin{tabular}{llllllllllllllll|l}
\hline
\textbf{Scan} & 24   & 37   & 40   & 55   & 63   & 65   & 69   & 83   & 97   & 105  & 106  & 110  & 114  & 118  & 122  & Mean \\  
\hline
\hline
COLMAP \cite{schonberger2016structure} & \textbf{0.9} & 2.89 & 1.63 & 1.08 & 2.18 & 1.94 & 1.61 & 1.3  & 2.34 & 1.28 & 1.1  & 1.42 & 0.76 & 1.17 & 1.14 & 1.52 \\  
MVSNet \cite{yao2018mvsnet} & 1.05 & 2.52 & 1.71 & 1.04 & 1.45 & 1.52 & 0.88 & 1.29 & 1.38 & 1.05 & 0.91 & 0.66 & 0.61 & 1.08 & 1.16 & 1.22 \\  
CasMVSNet \cite{gu2020cascade} & 1.03 & 2.55 & 1.60 & 0.92 & 1.51 & 1.67 & 0.81 & 1.43 & 1.22 & 1.05 & 0.83 & 0.67 & 0.54 & 0.94 & 1.10 & 1.19 \\
\hline
VolSDF \cite{yariv2021volume} & 4.03 & 4.21 & 6.12 & 0.91 & 8.24 & 1.73 & 2.74 & 1.82 & 5.14 & 3.09 & 2.08 & 4.81 & 0.6  & 3.51 & 2.18 & 3.41 \\  
NeuS \cite{wang2021neus} & 4.57 & 4.49 & 3.97 & 4.32 & 4.63 & 1.95 & 4.68 & 3.83 & 4.15 & 2.5  & 1.52 & 6.47 & 1.26 & 5.57 & 6.11 & 4.00 \\  
3DGS \cite{kerbl20233d} & 3.01 & 2.71 & 2.11 & 1.96 & 4.07 & 3.35 & 2.48 & 3.4 & 2.74 & 2.91 & 3.31 & 3.29 & 1.73 & 2.73 & 2.51 & 2.82 \\  
2DGS \cite{huang20242d} & 3.12 & \textbf{2.12} & 2.18 & 1.41 & 3.74 & 2.39 & 2.18 & 2.84 & 2.87 & 1.93 & 3.71 & 3.88 & 1.07 & 3.02 & 2.01 & 2.56 \\  
MASt3R \cite{leroy2024grounding} & 1.62 & 3.06 & 3.74 & 1.50 & 1.75 & 2.70 & 1.65 & 1.82 & 2.64 & 1.80 & 1.52 & 1.43 & 1.20 & 1.52 & 1.37 & 1.95 \\  
IBRNet-ft \cite{wang2021ibrnet} & 1.67 & 2.97 & 2.26 & 1.56 & 2.52 & 2.30 & 1.50 & 2.05 & 2.02 & 1.73 & 1.66 & 1.63 & 1.17 & 1.84 & 1.61 & 1.90 \\  
SparseNeuS-ft \cite{long2022sparseneus} & 1.29 & 2.27 & 1.57 & \underline{0.88} & 1.61 & 1.86 & 1.06 & 1.27 & 1.42 & 1.07 & 0.99 & 0.87 & \underline{0.54} & 1.15 & 1.18 & 1.27 \\  
\hline
IBRNet \cite{wang2021ibrnet} & 2.29 & 3.70 & 2.66 & 1.83 & 3.02 & 2.83 & 1.77 & 2.28 & 2.73 & 1.96 & 1.87 & 2.13 & 1.58 & 2.05 & 2.09 & 2.32 \\  
GeoNeRF \cite{johari2022geonerf} & 3.40 & 4.37 & 3.99 & 2.94 & 5.08 & 4.50 & 3.42 & 4.68 & 4.54 & 4.05 & 3.47 & 3.23 & 3.34 & 3.57 & 3.63 & 3.88 \\ 
SparseNeuS \cite{long2022sparseneus} & 1.68 & 3.06 & 2.25 & 1.1  & 2.37 & 2.18 & 1.28 & 1.47 & 1.8  & 1.23 & 1.19 & 1.17 & 0.75 & 1.56 & 1.55 & 1.64 \\ 
VolRecon \cite{ren2023volrecon} & 1.2  & 2.59 & 1.56 & 1.08 & 1.43 & 1.92 & 1.11 & 1.48 & 1.42 & 1.05 & 1.19 & 1.38 & 0.74 & 1.23 & 1.27 & 1.38 \\  
ReTR \cite{liang2024retr} & 1.05 & 2.31 & \underline{1.44} & 0.98 & 1.18 & 1.52 & 0.88 & 1.35 & 1.3  & 0.87 & 1.07 & 0.77 & 0.59 & 1.05 & 1.12 & 1.17 \\
GeoTransfer \cite{jena2024geotransfer} & 0.95 & \underline{2.23} & 1.45 & 0.94 & 1.26 & 1.67 & 0.81 & \underline{1.21} & 1.34 & 1.02 & \underline{0.84} & \underline{0.6} & 0.58 & 0.94 & \underline{1.02} & 1.12 \\ 
UfoRecon$^{\ast}$\footnotemark[1] \cite{ren2023volrecon} & \underline{0.91}  & 2.32 & \textbf{1.41} & 0.89 & \underline{1.17} & \textbf{1.21} & \underline{0.74} & 1.24 & \textbf{1.18} & \textbf{0.83} & \textbf{0.77} & 0.64 & 0.56 & \underline{0.89} & \underline{1.02} & \underline{1.05} \\  
\hline
Ours & 0.97 & 2.28 & 1.5 & \textbf{0.85} & \textbf{1.14} & \underline{1.26} & \textbf{0.73} & \textbf{1.19} & \underline{1.2} & \underline{0.86} & \textbf{0.77} & \textbf{0.58} & \textbf{0.48} & \textbf{0.84} & \textbf{0.95} & \textbf{1.04} \\  
\hline
\end{tabular}
}
\caption{Quantitative comparison on the DTU dataset \cite{aanaes2016large}. Best and second best methods are \textbf{emboldened} and \underline{underlined} respectively. UfoRecon$^{\ast}$ \cite{ren2023volrecon} refers to the results generated with pretrained model provided in their code.\\ (Inference time: Ours 0.8s, UfoRecon\cite{ren2023volrecon} 66s)} \label{tab:table1}
\vspace{-7pt}
\end{table*}

In this section, we showcase the performance of our proposed approach. Firstly, we offer an overview of our experimental configurations, implementation specifics, datasets, and baseline methods. Secondly, we present both quantitative and qualitative comparisons on three commonly used multi-view datasets, namely DTU \cite{aanaes2016large}, BlendedMVS \cite{yao2020blendedmvs} and Tanks and Temples \cite{knapitsch2017tanks}. Finally, we perform ablation studies to evaluate the impact of the components of our proposed method.
\vspace{-15pt}
\paragraph{Datasets}
In line with prior research \cite{long2022sparseneus, ren2023volrecon, liang2024retr, na2024uforecon}, we employ the DTU dataset \cite{aanaes2016large} for the training phase. The DTU dataset \cite{aanaes2016large} is characterized by indoor multi-view stereo data, featuring ground truth point clouds from 124 distinct scenes and under 7 different lighting conditions. Throughout our experiments, we utilize the same set of 15 scenes as \cite{long2022sparseneus, ren2023volrecon, liang2024retr, jena2024geotransfer, na2024uforecon} for testing purposes, reserving the remaining scenes for training. Concerning the BlendedMVS dataset \cite{yao2020blendedmvs}, we opt for the same scenes as used in SparseNeuS \cite{long2022sparseneus}, with some additional ones chosen randomly. For each scene in either DTU \cite{aanaes2016large} or BMVS \cite{yao2020blendedmvs}, we use the same $3$ sparse input views following SparseNeuS \cite{long2022sparseneus}. To ensure impartial evaluation on DTU \cite{aanaes2016large}, we use the foreground masks from IDR \cite{yariv2020multiview} to mask the reconstructed meshes and evaluate how well our approach performs on the test set, consistent with prior research \cite{long2022sparseneus, ren2023volrecon, liang2024retr, jena2024geotransfer, na2024uforecon}. Additionally, to assess the generalization capability of our proposed framework, we qualitatively compare our method on the BlendedMVS dataset \cite{yao2020blendedmvs} and provide some additional examples om the Tanks and Temples dataset \cite{knapitsch2017tanks} without any fine-tuning. For our novel-view synthesis experiments, we follow the same split of testing images within a scene as well as the same testing scenes as in MVSGaussian \cite{liu2024fast}. We also include some additional reconstruction and novel view synthesis qualitative comparisons in the supplementary material, including some video examples.


\vspace{-15pt}
\paragraph{Baselines}
In order to showcase the efficacy of our method, we conducted comparisons with a) SparseNeus \cite{long2022sparseneus}, VolRecon \cite{ren2023volrecon}, ReTR \cite{liang2024retr}, GeoTransfer \cite{jena2024geotransfer} and UfoRecon \cite{na2024uforecon}, the leading generalizable neural surface reconstruction approaches; b) Generalizable neural/$3$DGS rendering methods PixelNeRF \cite{yu2021pixelnerf}, IBRNet \cite{wang2021ibrnet}, MVSNeRF \cite{chen2021mvsnerf}, GeoNeRF \cite{johari2022geonerf}, and MVSGaussian \cite{liu2024fast}; c) Neural implicit reconstruction methods VolSDF \cite{yariv2021volume}, NeuS \cite{wang2021neus} and Gaussian Splatting based methods 3D Gaussian Splatting \cite{kerbl20233d} and 2D Gaussian Splatting \cite{huang20242d} that necessitate scene-specific training; and finally, d) Well-established multi-view stereo (MVS) methods Colmap \cite{schonberger2016structure}, MVSNet \cite{yao2018mvsnet}, and CasMVSNet \cite{gu2020cascade} as well as MASt3R \cite{leroy2024grounding}. 
\vspace{-15pt}
\paragraph{Implementation details}
Our model is implemented using PyTorch \cite{paszke2019pytorch}. During training, we utilize an image resolution of $640 \times 512$, with $N$ (the number of source images) set to $3$. Network architecture follows mostly \cite{liu2024fast}. Training extends over $300$k steps using the Adam optimizer \cite{kingma2014adam} on a single RTX A$6000$ GPU, with an initial learning rate of $5 \times 10^{-4}$. To extract meshes from the reconstructed $2$D splats, we render depth maps of the training views. We then utilize truncated signed distance function (TSDF) fusion to combine the reconstructed depth maps. During TSDF fusion, we set the voxel size to $1.5$ following \cite{long2022sparseneus, ren2023volrecon, liang2024retr, jena2024geotransfer, na2024uforecon}.

\subsection{Sparse View Reconstruction on DTU}
\begin{figure*}
    \centering
    \includegraphics[width=1.0\linewidth]{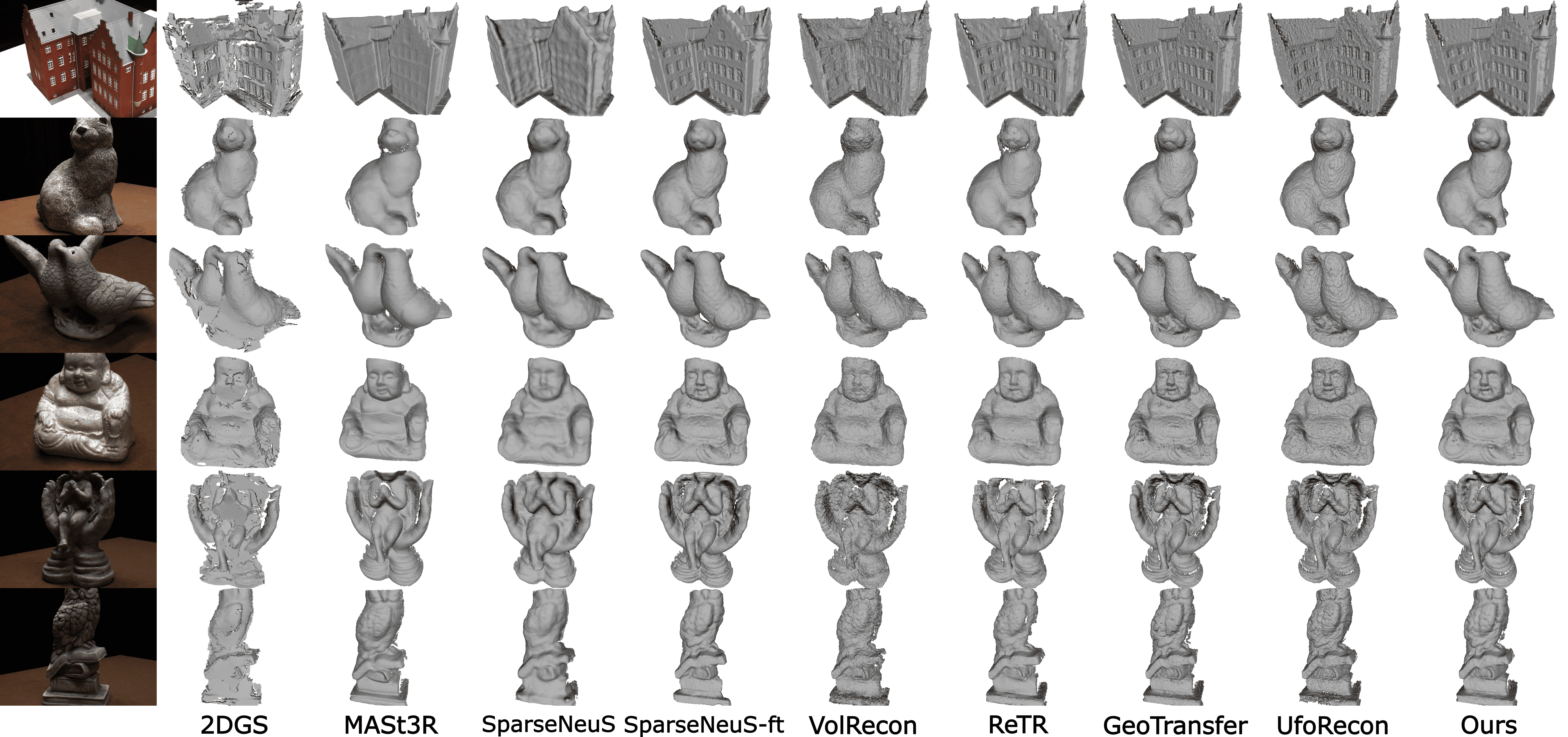}
    \vspace{-15pt}
    \caption{Qualitative comparison of reconstructions from 3 input views in datatset DTU. (Inference time: Ours 0.8s, UfoRecon\cite{ren2023volrecon} 66s).}
    \vspace{-8pt}
    \label{fig:DTU}
\end{figure*}

\vspace{-5pt}
We conduct surface reconstruction with sparse views (only $3$ views) on the DTU dataset \cite{aanaes2016large} and assess the predicted surface by comparing it to the ground-truth point clouds using the chamfer distance metric. To facilitate a fair comparison, we followed the evaluation process employed in \cite{long2022sparseneus, ren2023volrecon, liang2024retr, jena2024geotransfer, na2024uforecon} and adhered to the same testing split as described in them. As indicated in Tab. \ref{tab:table1}, we offer superior results compared to neural implicit \cite{yariv2021volume, wang2021neus} and Gaussian Splatting based \cite{kerbl20233d, huang20242d} methods that use scene-specific training, but struggle in sparse input-view setup to lack of prior knowledge. We clearly outperform MASt3R \cite{leroy2024grounding} as well, where the mesh reconstructions for MASt3R have been obtained by performing TSDF fusion (with the same voxel size of $1.5$) on the depth maps inferred by MASt3R. Additionally, our method (only DTU trained) surpasses almost all the generalizable implicit methods \cite{long2022sparseneus, ren2023volrecon, liang2024retr} by a good margin, \ie by $36.58\%$, $32.69\%$ and $12.5\%$ respectively. Furthermore, our approach exhibits superior performance compared to well-known multi-view stereo (MVS) methods like Colmap \cite{schonberger2016structure}, MVSNet \cite{yao2018mvsnet} and CasMVSNet \cite{gu2020cascade}. Our approach also demonstrates superior performance in comparison to GeoTransfer \cite{jena2024geotransfer} (by $7.14\%$) and narrowly outperforms UfoRecon \cite{na2024uforecon}, which are the latest state-of-the-art generalizable neural implicit reconstruction methods. Here, we report results for UfoRecon \cite{na2024uforecon} by averaging chamfer metrics over multiple runs using their code and pretrained model\footnotemark[1]\footnotetext[1]{\href{https://github.com/Youngju-Na/UFORecon}{https://github.com/Youngju-Na/UFORecon}}. Despite our best efforts, we could not reproduce the chamfer metrics originally reported in their paper. Additionally, we present qualitative results of sparse view reconstruction in Fig.\ref{fig:DTU}, showcasing that our reconstructed geometry features more expressive and detailed surfaces compared to the current state-of-the-art methods.


\subsection{Generalization on BlendedMVS and Tanks and Temples} 
\begin{figure}
    \centering
    \includegraphics[width=1.0\linewidth]{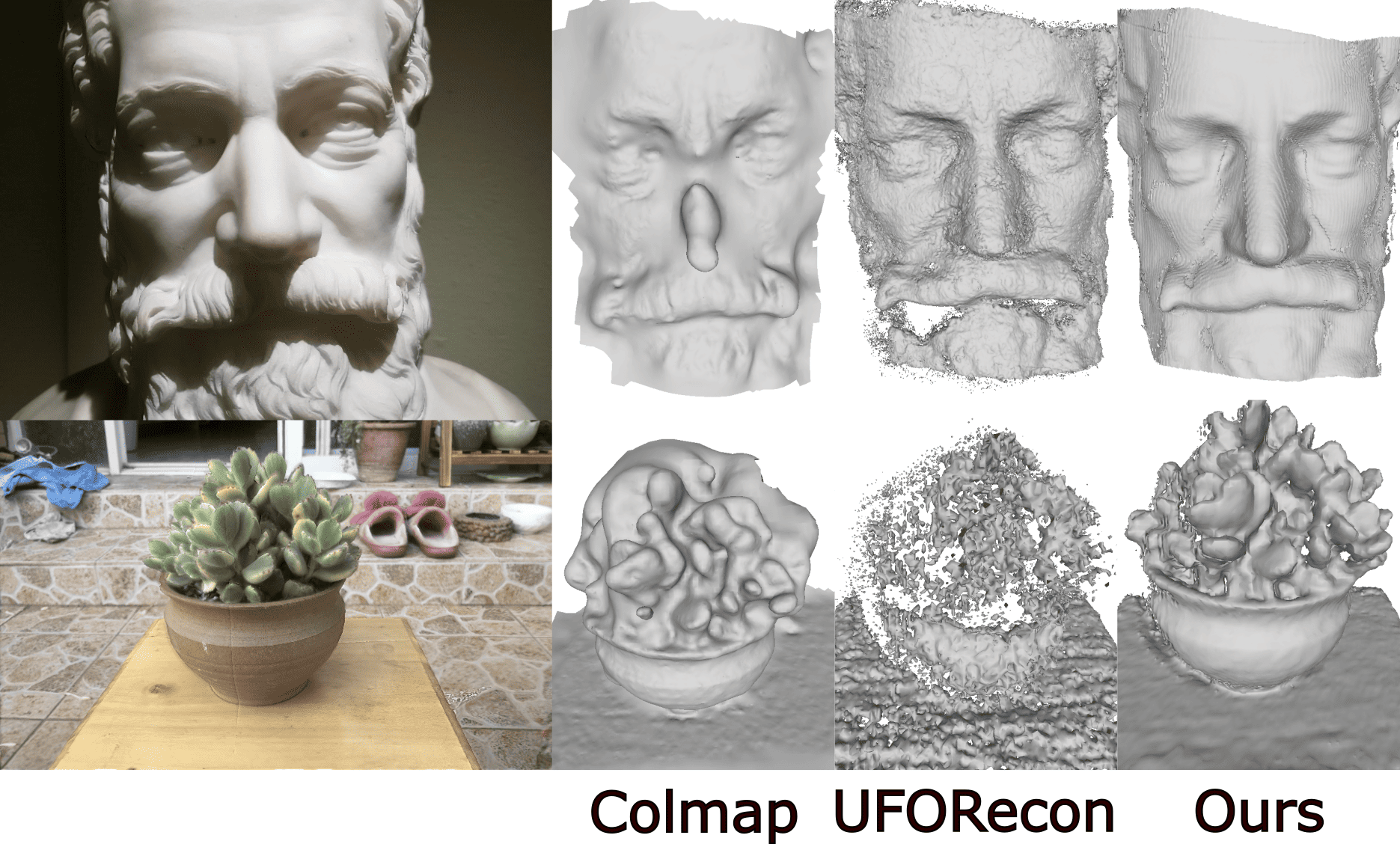}
    \\
    \vspace{5pt}
    \includegraphics[width=0.88\linewidth]{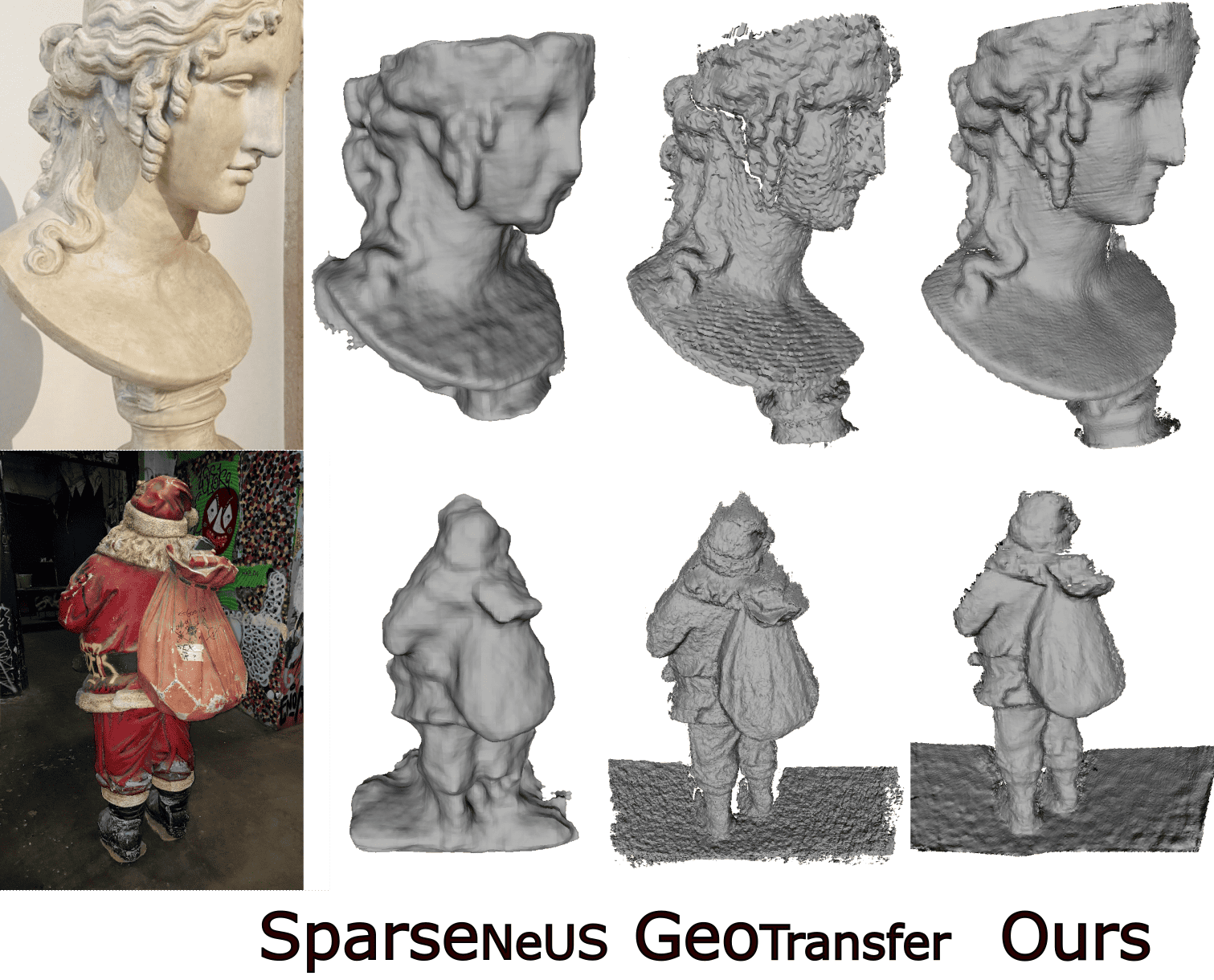}
    \caption{Qualitative comparison of reconstructions from 3 input views in datatset BMVS. Note that we reconstruct detailed surfaces with our method without any fine-tuning.}
    \vspace{-10pt}
    \label{fig:BMVS}
\end{figure}

\vspace{-5pt}
To demonstrate the generalization prowess of our proposed approach, we perform additional evaluations on the BlendedMVS dataset \cite{yao2020blendedmvs} without resorting to any fine-tuning. The high-fidelity reconstructions of scenes and objects, as illustrated in Fig. \ref{fig:BMVS}, affirms the efficacy of our method in terms of its generalization capabilities. Our method is able to obtain more detailed surfaces in comparison to Colmap \cite{schonberger2016structure}, SparseNeuS \cite{long2022sparseneus}, GeoTransfer \cite{jena2024geotransfer} as well as the recent SOTA UfoRecon \cite{na2024uforecon}.

\begin{figure}
    \centering
    \includegraphics[width=1.0\linewidth]
    {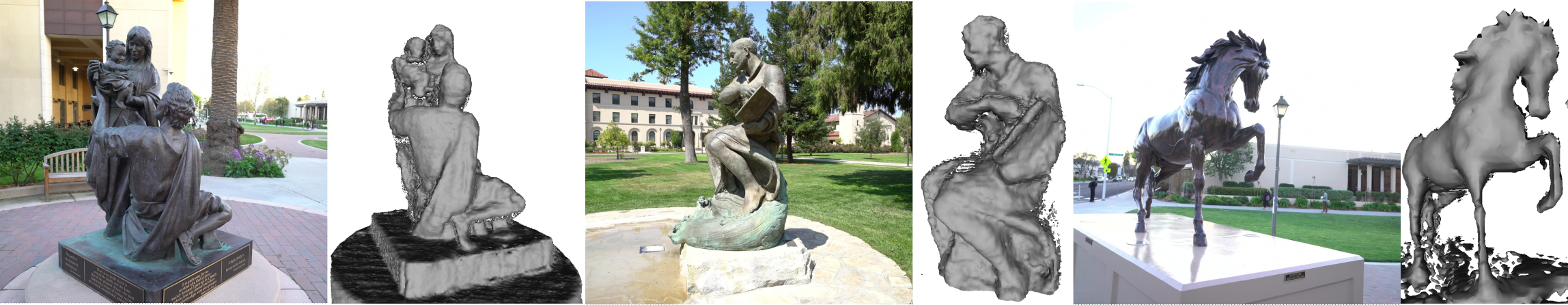}
    \caption{Qualitative results of reconstructions from 3 input views in datatset Tanks and Temples using our method without any fine-tuning. Notice that these are very challenging outdoor cases.}
    \vspace{-10pt}
    \label{fig:TNT}
\end{figure}

As shown in Fig. \ref{fig:TNT}, we further emphasize the generalization capabilities of our method by providing reconstruction examples on scenes of the Tanks and Temples dataset.

\begin{figure*}[t!]
    \centering
    \setlength{\tabcolsep}{1pt}
    \begin{minipage}{\linewidth}
    \centering
    \begin{tabular}{@{}cccccc@{}}
        \includegraphics[width=0.162\linewidth]{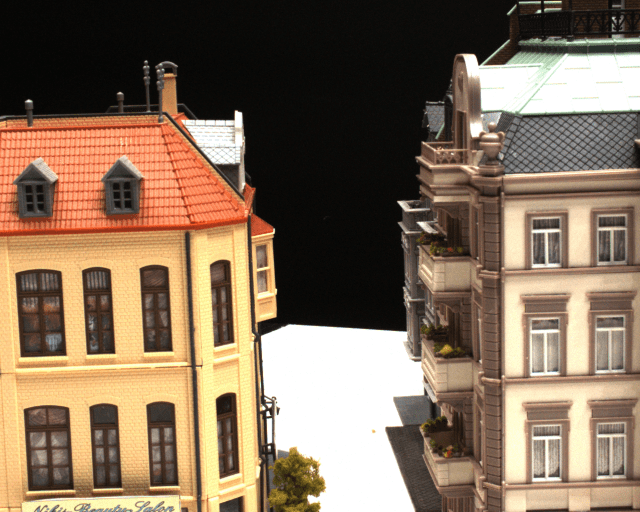} & 
        \includegraphics[width=0.162\linewidth]{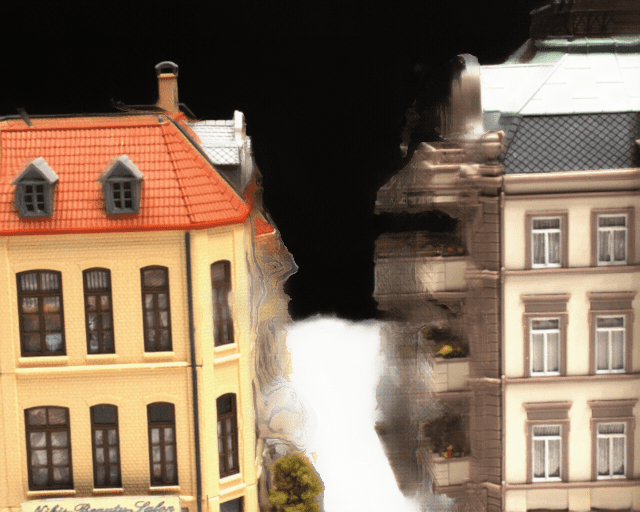} & 
        \includegraphics[width=0.162\linewidth]{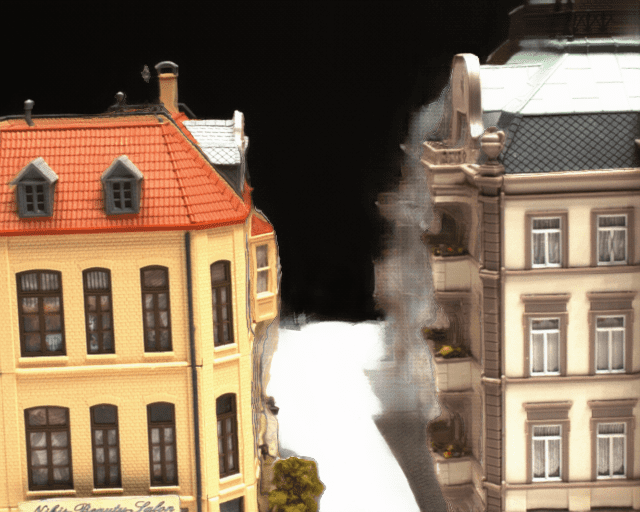} &
        \includegraphics[width=0.162\linewidth]{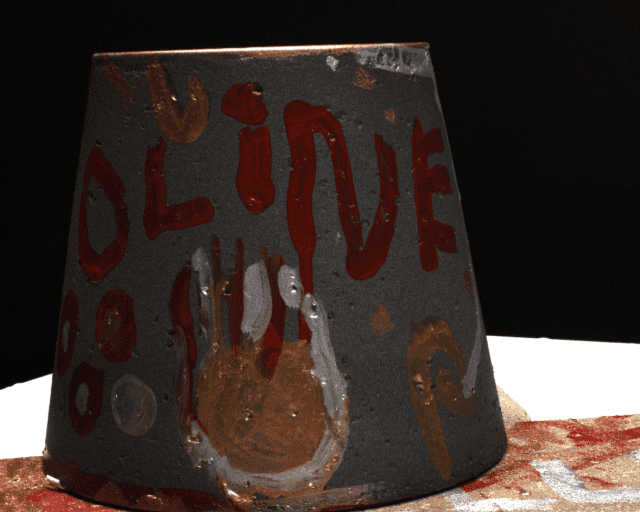} & 
        \includegraphics[width=0.162\linewidth]{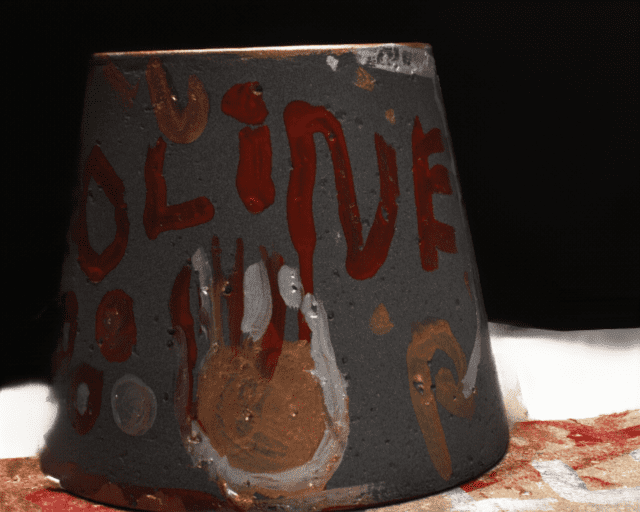} & 
        \includegraphics[width=0.162\linewidth]{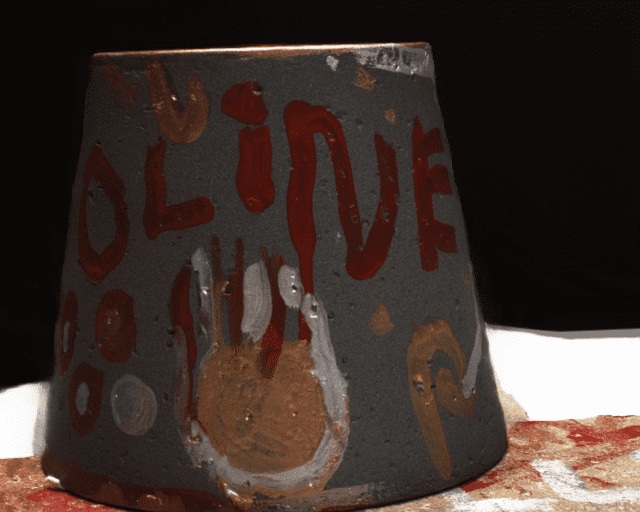} \\
        \includegraphics[width=0.162\linewidth]{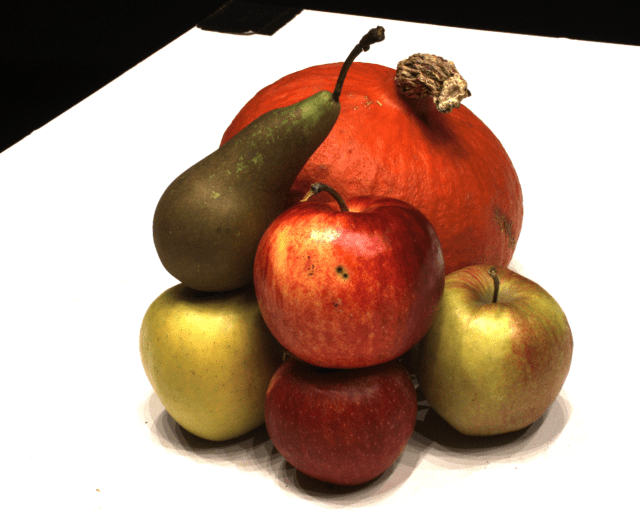} & 
        \includegraphics[width=0.162\linewidth]{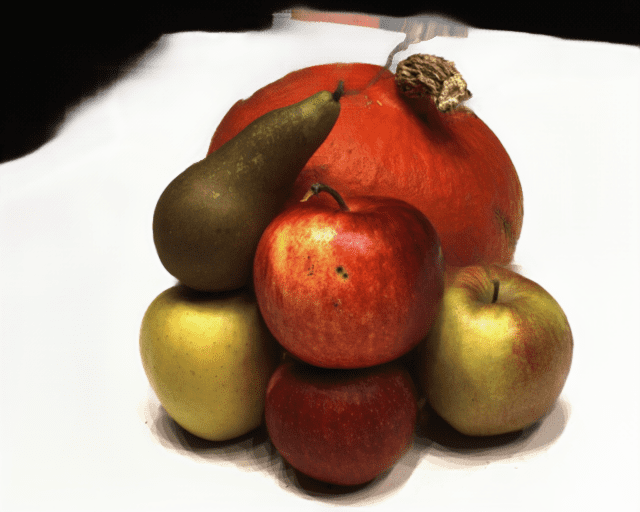} & 
        \includegraphics[width=0.162\linewidth]{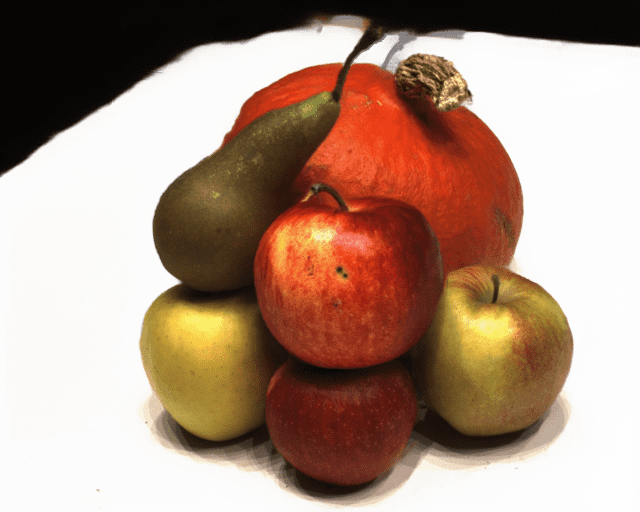} & 
        \includegraphics[width=0.162\linewidth]{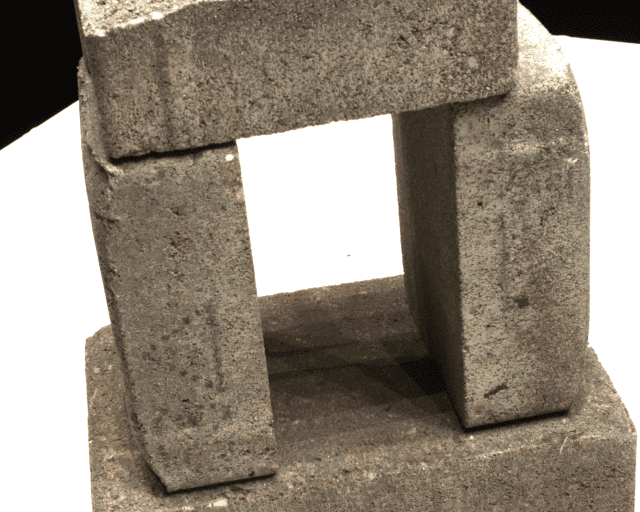} & 
        \includegraphics[width=0.162\linewidth]{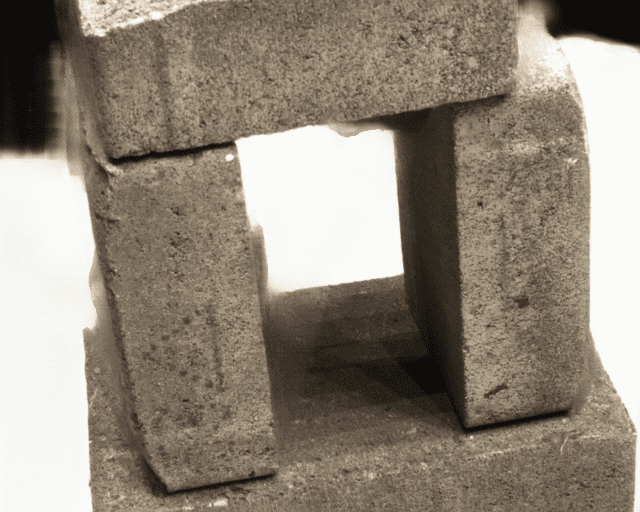} & 
        \includegraphics[width=0.162\linewidth]{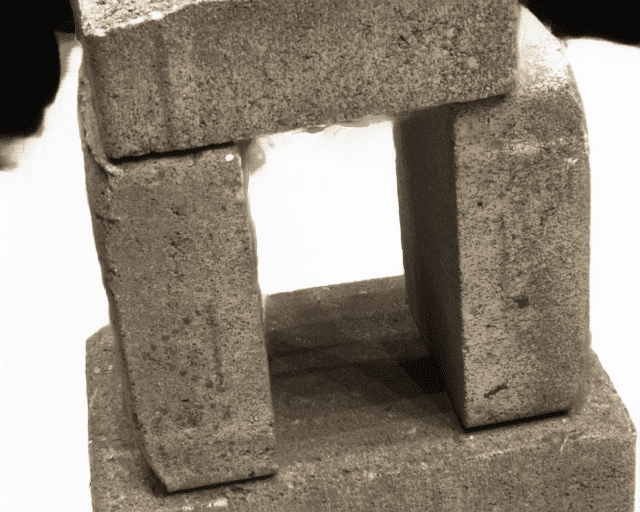} \\
        GT & MVSGaussian \cite{liu2024fast} & Ours & GT & MVSGaussian \cite{liu2024fast} & Ours \\
    \end{tabular}
    \end{minipage}
    \caption{Novel-View synthesis qualitative evalutation on DTU \cite{aanaes2016large} using $3$ source images. We outperform SOTA MVSGaussian \cite{liu2024fast} and exhibit sharper view synthesis results in regions with low input view overlap}
    \label{fig:nvs_grid}
\end{figure*}

\subsection{Novel-view synthesis on DTU}
\vspace{-5pt}
We evaluate novel-view synthesis results on the DTU \cite{aanaes2016large} dataset. We compare our method against 
MVSGaussian \cite{liu2024fast} qualitatively (as shown in Fig. \ref{fig:nvs_grid}) as well as other
recent methods quantitatively on the mean PSNR, SSIM and LPIPS metrics over all scenes by following the same data split for training and testing as them. We use $3$ input source images for all the methods concerned. Our results can be summarized in Tab.\ref{tab:dtu_nv}.
\begin{table}[t]
\centering
\setlength{\tabcolsep}{5pt}
\resizebox{0.8\linewidth}{!}{
\begin{tabular}{@{}lccc@{}}
\toprule
Method & PSNR $\uparrow$ & SSIM $\uparrow$ & LPIPS $\downarrow$ \\
\midrule
PixelNeRF~\cite{yu2021pixelnerf} & 19.31 & 0.789 & 0.382 \\
IBRNet~\cite{wang2021ibrnet} & 26.04 & 0.917 & 0.191 \\ 
MVSNeRF~\cite{chen2021mvsnerf}  & 26.63 & 0.931 & 0.168 \\
ENeRF~\cite{lin2022efficient}  & 27.61 & 0.957 & 0.089 \\
MatchNeRF~\cite{chen2023explicit} & 26.91 & 0.934 & 0.159 \\
MVSGaussian~\cite{liu2024fast} & \underline{28.21} & \textbf{0.963} & \underline{0.076} \\
Ours & \textbf{28.33} & \underline{0.938} & \textbf{0.073} \\
\bottomrule
\end{tabular}
}
\caption{Quantitative results of NVS generalization on the DTU test set. The best result is \textbf{bold}, and the second-best is \underline{underlined}.}
\label{tab:dtu_nv}
\end{table}
This experiment illustrates that robust features of the $3$D foundation model MASt3R aid in improving novel view synthesis performance. Overall, our method offers state-of-the-art reconstruction while also improving upon the novel-view synthesis performance to become the new the state-of-the-art in NVS in the sparse DTU setting introduced by SparseNeus\cite{long2022sparseneus}. Hence, as opposed to other implicit reconstruction methods \cite{long2022sparseneus, ren2023volrecon, liang2024retr} (which have relatively poor novel view synthesis capabilities as detailed in \cite{jena2024geotransfer}), we offer a dual advantage in both the reconstruction and novel-view synthesis results.

\subsection{Ablation studies}
\label{sec:abl}
\vspace{-5pt}
We perform the following ablation studies in the full training scenario using all testing scenes of the DTU dataset as evaluation. 


\vspace{-15pt}
\paragraph{\textbf{Impact of splatting method}} We ablate the impact of each splatting method as well as the $\mathcal{L}_{\text{depth}}$ depth loss (Eq.\ref{eq:loss}). Without 2DGS \cite{huang20242d}, (\ie using 3DGS as in MVSGaussian \cite{liu2024fast}) the resulting depth has inconsistencies across views. This results in TSDF producing incorrect fusions, with surfaces that are warped/disconnected, due to which the method fails to obtain conherent surfaces. For depth loss $\mathcal{L}_{\text{depth}}$, guiding the mean 2DGS splatted depth with the ground truth depth supervision of DTU dataset \cite{aanaes2016large} has a major impact on the reconstruction quality, improving mean chamfer distance by $71.8\%$. The mean chamfer changes from $4.37$ across all test scenes over the two sets of views on DTU to our final mean chamfer of $1.04$ for the model that uses MASt3R features as input. 

\vspace{-15pt}
\paragraph{\textbf{Impact of external features}} In tab.\ref{tab:feats}, we ablate the impact of using features from $3$D foundation models such as MASt3R \cite{leroy2024grounding} over $2$D ones such as DinoV2 \cite{oquab2023dinov2} which lack multi-view consistency. We see that compared to just using image features extracted using a feature pyramid network (FPN), concatenating DinoV2 \cite{oquab2023dinov2} gives us a boost of only $0.85\%$. However, when using features from the dense local feature head of MASt3R \cite{leroy2024grounding}, we get a boost of $11.11\%$, which shows that features from $3$D foundation models are well suited for MVS based tasks compared to DinoV2 \cite{oquab2023dinov2}. 

\begin{table}[h]
\centering
{
\hspace*{1.2\leftmargin}\begin{tabular}{ cc}
 \hline
 Features & Mean CD$\downarrow$\\
 \hline
 Image feats & 1.17 \\
 Img+DinoV2 \cite{oquab2023dinov2} feats & 1.16\\
 Ours (Img+MASt3R \cite{leroy2024grounding} feats) & \textbf{1.04}\\
 \hline
\end{tabular}
}
\caption{Impact of using different features on $3$D reconstruction.} \label{tab:feats}
\end{table}



\subsection{Inference time}
Once we infer pixel-aligned 2DGS parameters in a feed-forward manner, we can leverage them to render our depth maps significantly faster than previous generalizable implicit methods for reconstruction. We provide here inference speeds for ours and main baselines VolRecon \cite{ren2023volrecon}, ReTR \cite{liang2024retr}, GeoTransfer \cite{jena2024geotransfer} and UfoRecon \cite{na2024uforecon}. Depth map inference times is about $30$s as reported in their respective supplementary sections, while for UfoRecon \cite{na2024uforecon} it is higher at around $60$s. We reproduced this on a RTX A$6000$ and we obtained $0.88$s for ours, against $32$s for GeoTransfer \cite{jena2024geotransfer}, $31$s for VolRecon \cite{ren2023volrecon}, $37$s for ReTR \cite{liang2024retr}, and $66$s for UfoRecon \cite{na2024uforecon}.

\section{Conclusion}
\label{sec:conclusion}
\vspace{-5pt}
We proposed the first generalizable $2$D Gaussian splatting based approach, enabling fast multi-view reconstruction with a speedup factor of nearly $2$ orders of magnitude compared to implicit generalizable SOTA methods. Our approach benefits from the input of deep features extracted from existing $2$D and $3$D foundation models, namely monocular semantic features from DinoV2 \cite{oquab2023dinov2} and pairwise image features from MASt3R \cite{leroy2024grounding} which encodes dense correspondences between input images. We also achieve state-of-the-art results on the DTU dataset \cite{aanaes2016large} in both of our tasks of novel view synthesis and $3$D reconstruction, with strong generalization on the BlendedMVS dataset \cite{yao2020blendedmvs}, and promising results on the more challenging Tanks and Temples dataset \cite{knapitsch2017tanks}.

\clearpage
\appendix
\renewcommand{\thesection}{\Alph{section}}

\twocolumn[{
\vspace{2ex}
\begin{center}
    {\Large \bfseries SparSplat: Fast Multi-View Reconstruction with Generalizable 2D Gaussian Splatting}\\[0.5ex]
    {\large \bfseries – Supplementary Material –}\\[1ex]
    {\normalsize
    Shubhendu Jena\textsuperscript{1}, Shishir Reddy Vutukur\textsuperscript{2}, Adnane Boukhayma\textsuperscript{1}\\
    \textsuperscript{1}Inria, Univ. Rennes, CNRS, IRISA\\
    \textsuperscript{2}Technical University of Munich}
\end{center}
\vspace{2ex}
}]

Here, we show qualitative video comparisons of our reconstruction results to other methods to visually demonstrate the impact of our approach. Additional qualitative results for our novel view synthesis results are also included, and finally we conclude with some additional experimental details on our evaluation datasets of DTU \cite{aanaes2016large} and BlendedMVS \cite{yao2020blendedmvs} as well as additional implementation details.
\\
\noindent \textbf{Implementation Details.}
Following~\cite{liu2024fast}, for depth estimation, we sample $64$ and $8$ depth planes for the coarse and fine stages, respectively.
We set $\lambda_s=0.1$, $\lambda_p=0.05$, $\lambda_\alpha=0.05$, $\lambda_\beta=0.05$ and $\lambda_\gamma=0.05$, $\lambda^1=0.5$ and $\lambda^2=1$ for our loss weights.

\section{Additional qualitative comparison on 3D reconstruction}
Based on our $3$D reconstruction experiments, we have included some additional video visualizations of our surface reconstructions in the included supplementary material. There's $1$ on DTU \cite{aanaes2016large}, namely DTU\_Scan$122$.mp4 and $1$ on BlendedMVS \cite{yao2020blendedmvs}, namely BMVS\_Scan$2$.mp4. We also have a video result for the novel view synthesis task, namely NVS\_scan$1$.mp4

\section{Additional qualitative comparison on Novel View Synthesis}
\begin{figure*}[t]
    \centering
    \setlength{\tabcolsep}{1pt}
    \begin{minipage}{\linewidth}
    \centering
    \begin{tabular}{@{}cccccc@{}}
        \includegraphics[width=0.162\linewidth]{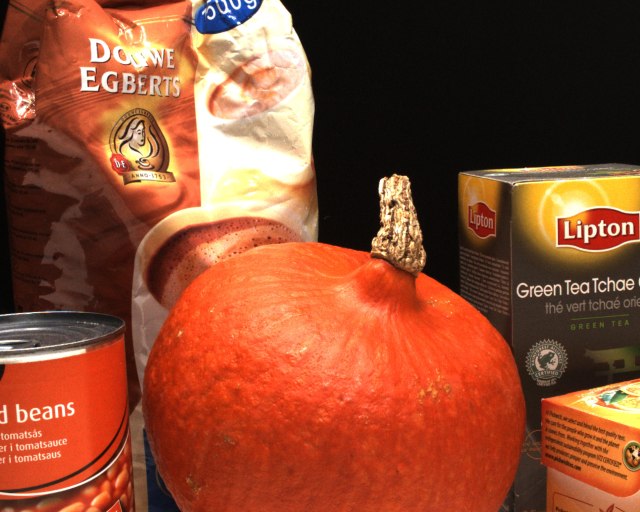} & 
        \includegraphics[width=0.162\linewidth]{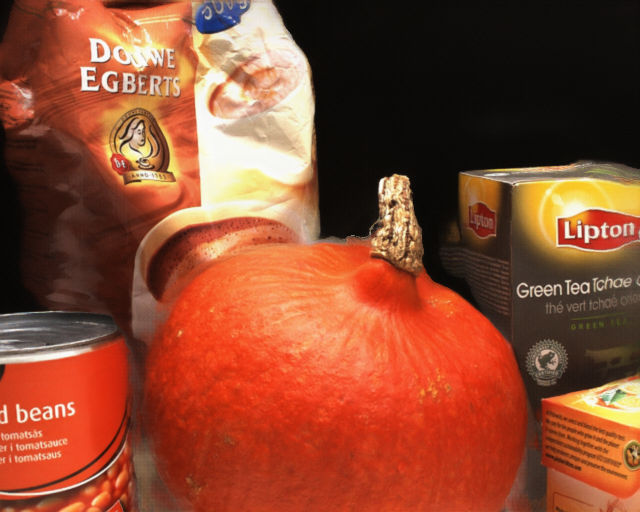} & 
        \includegraphics[width=0.162\linewidth]{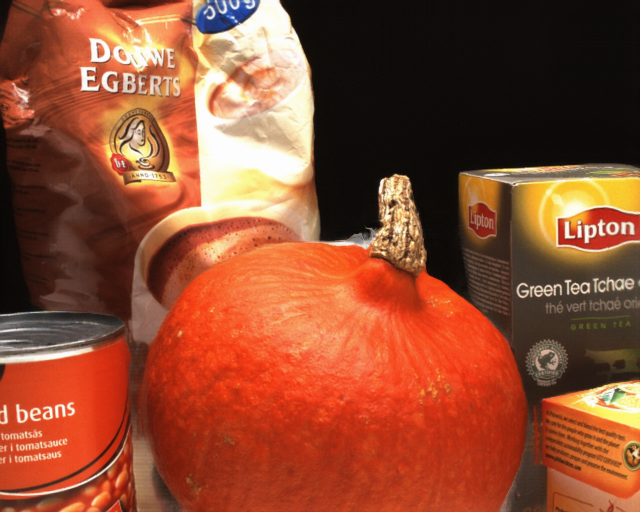} &
        \includegraphics[width=0.162\linewidth]{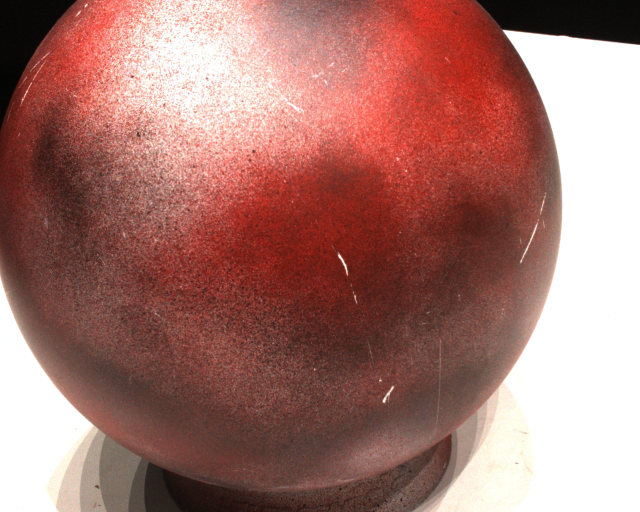} & 
        \includegraphics[width=0.162\linewidth]{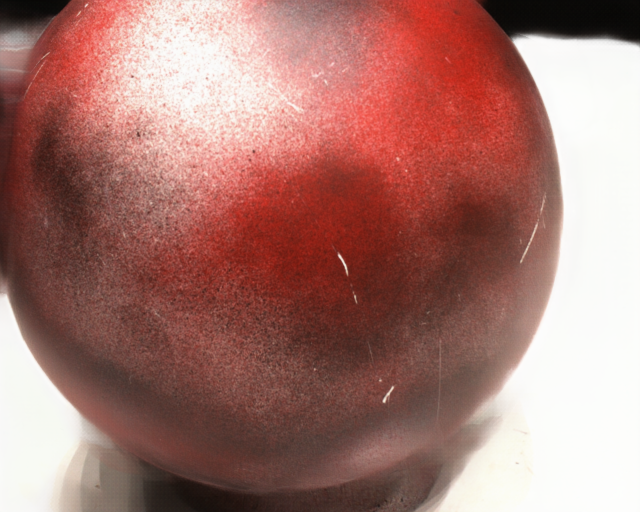} & 
        \includegraphics[width=0.162\linewidth]{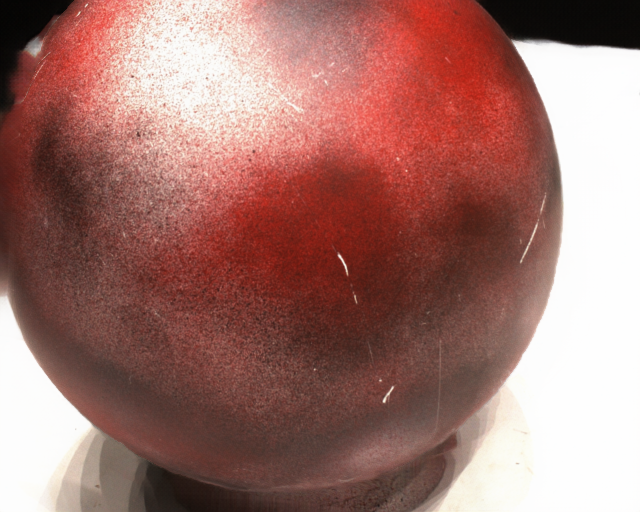} \\
        \includegraphics[width=0.162\linewidth]{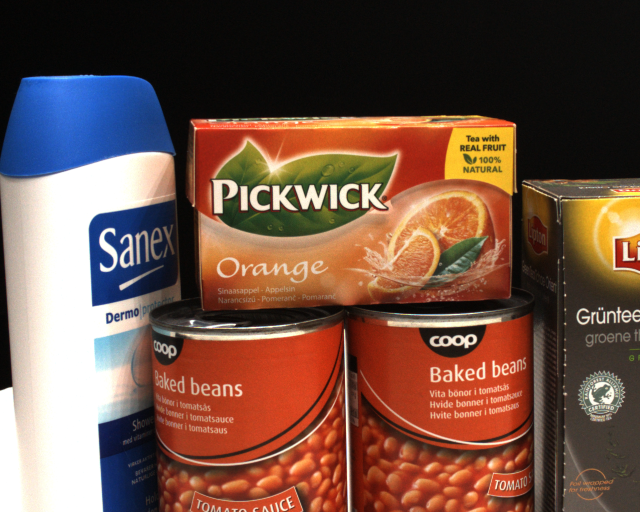} & 
        \includegraphics[width=0.162\linewidth]{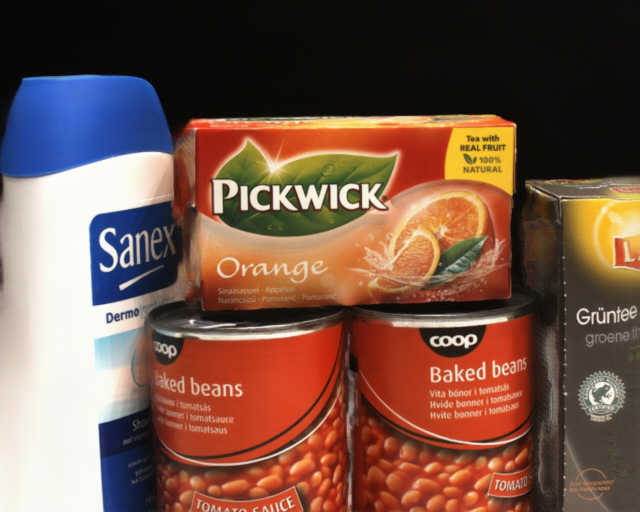} & 
        \includegraphics[width=0.162\linewidth]{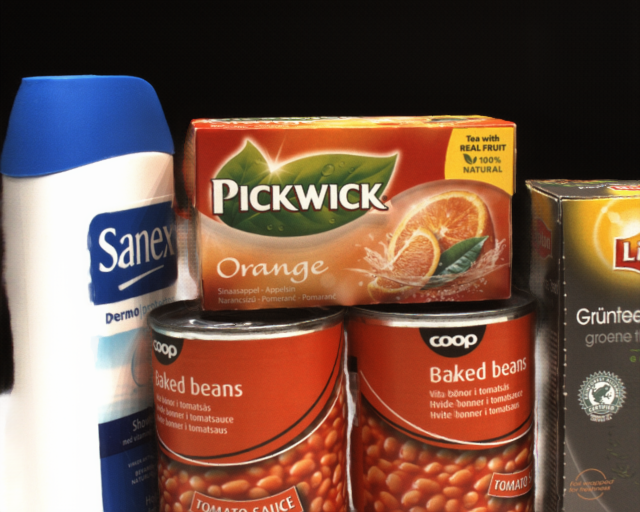} & 
        \includegraphics[width=0.162\linewidth]{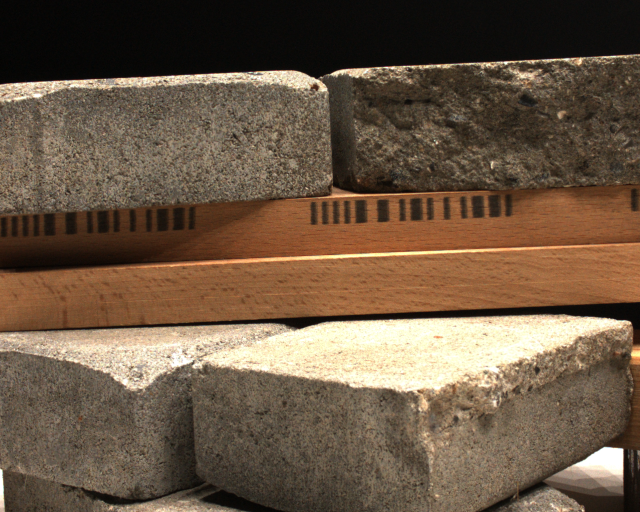} & 
        \includegraphics[width=0.162\linewidth]{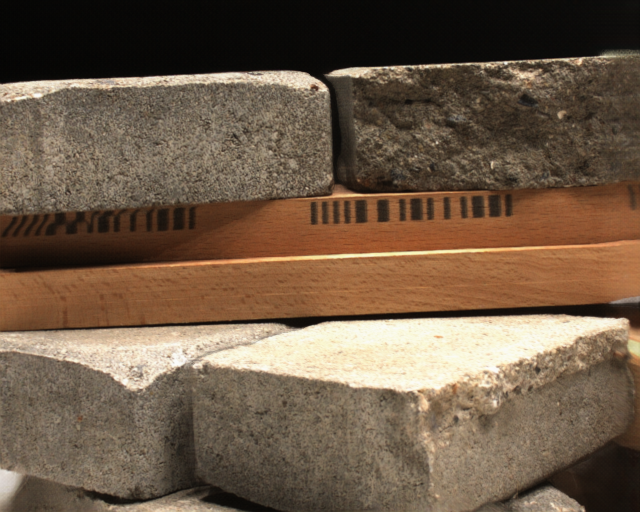} & 
        \includegraphics[width=0.162\linewidth]{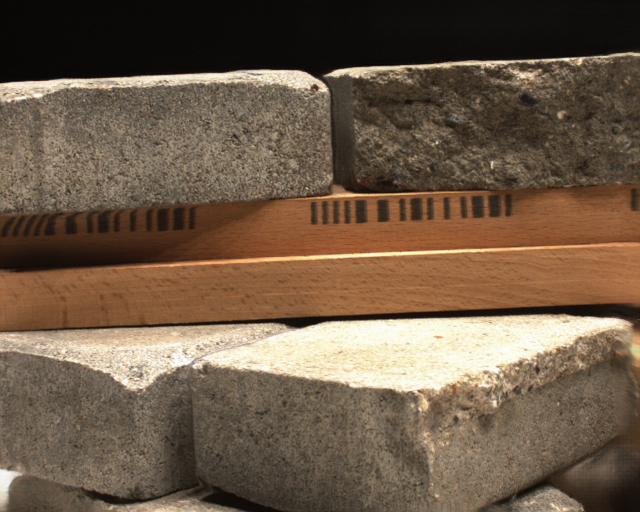} \\
        \includegraphics[width=0.162\linewidth]{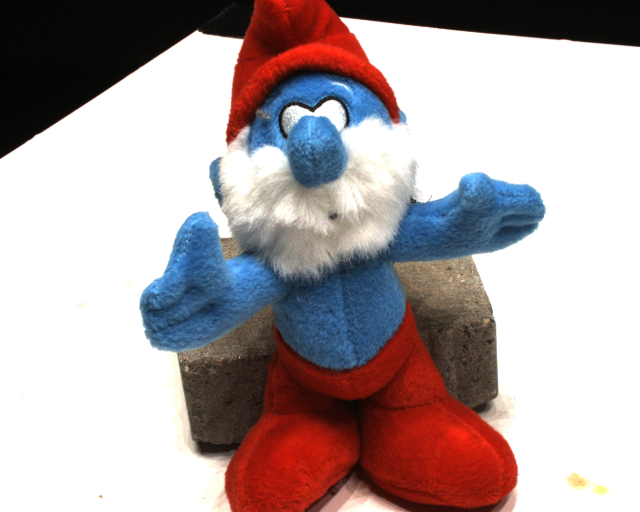} & 
        \includegraphics[width=0.162\linewidth]{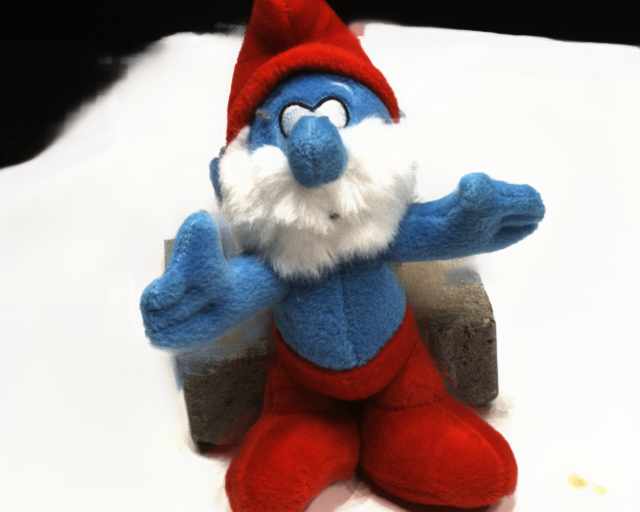} & 
        \includegraphics[width=0.162\linewidth]{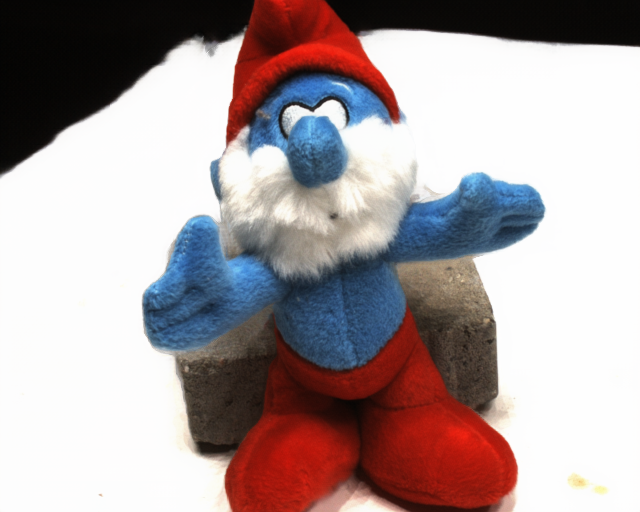} &
        \includegraphics[width=0.162\linewidth]{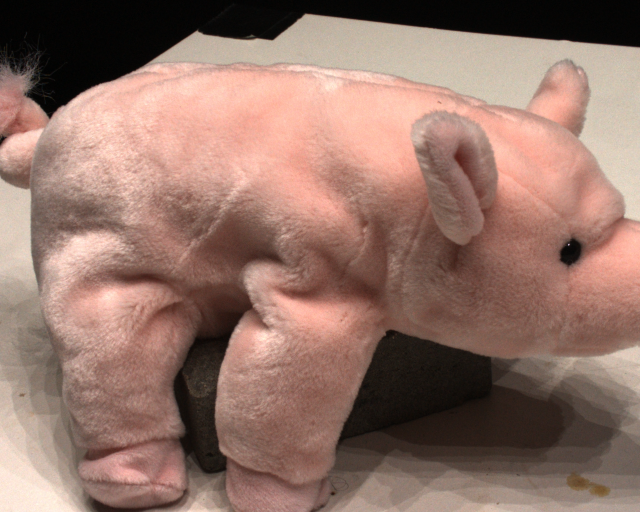} & 
        \includegraphics[width=0.162\linewidth]{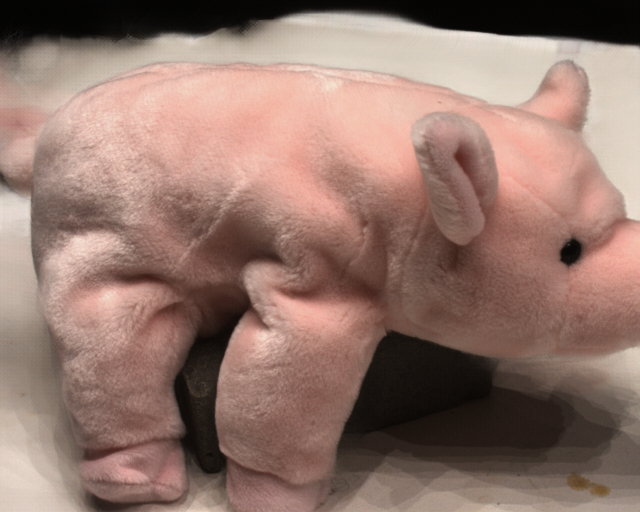} & 
        \includegraphics[width=0.162\linewidth]{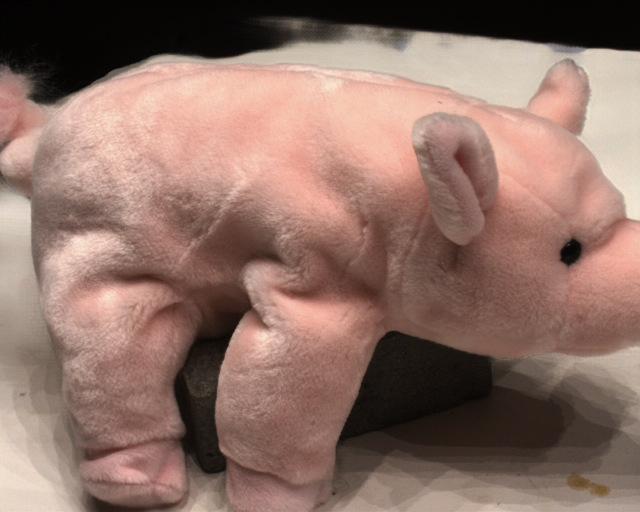} \\
        \includegraphics[width=0.162\linewidth]{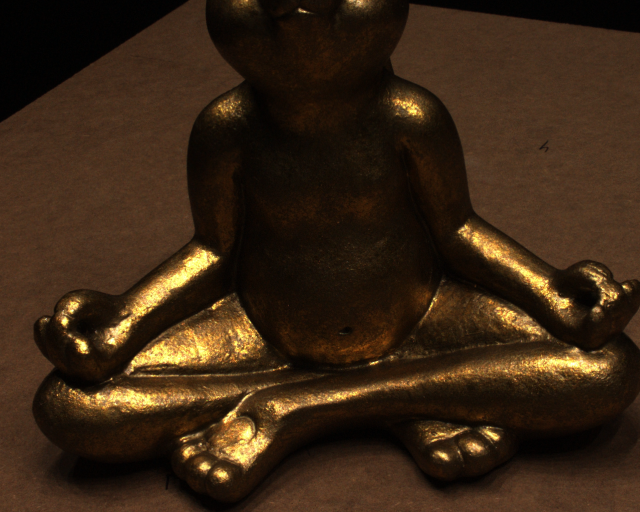} & 
        \includegraphics[width=0.162\linewidth]{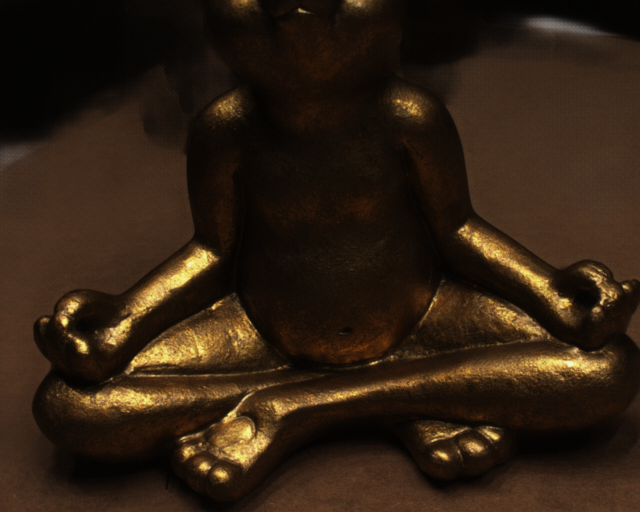} & 
        \includegraphics[width=0.162\linewidth]{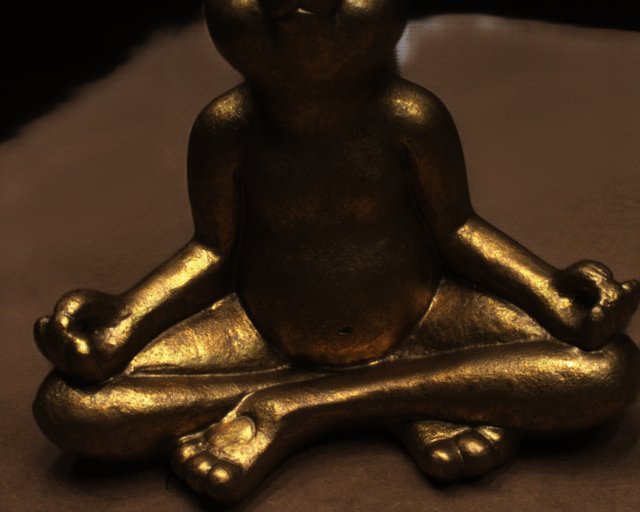} & 
        \includegraphics[width=0.162\linewidth]{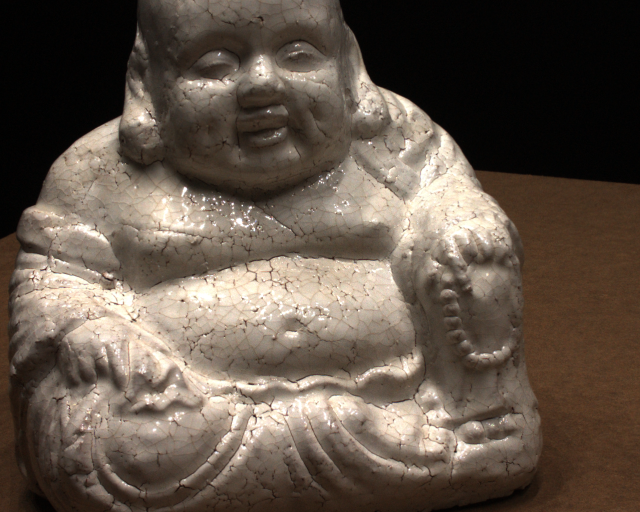} & 
        \includegraphics[width=0.162\linewidth]{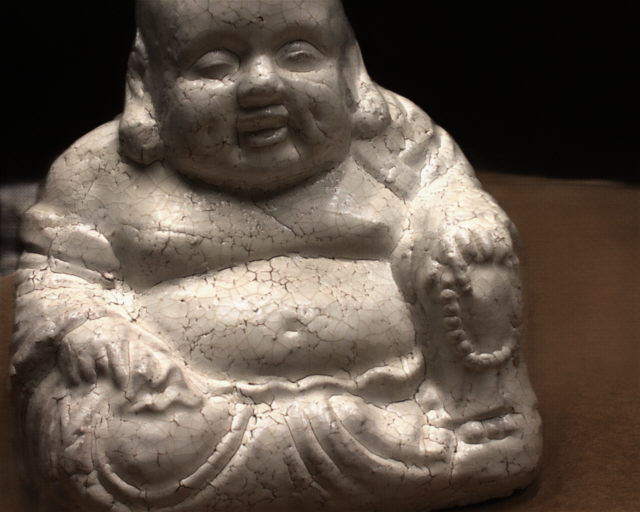} & 
        \includegraphics[width=0.162\linewidth]{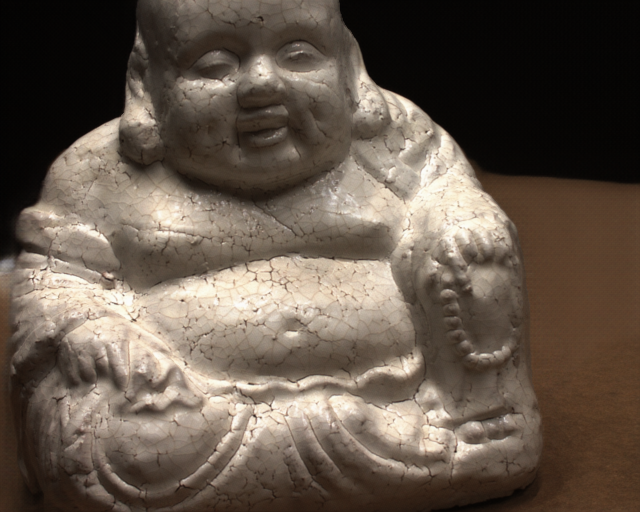} \\
        GT & MVSGaussian \cite{liu2024fast} & Ours & GT & MVSGaussian \cite{liu2024fast} & Ours \\
    \end{tabular}
    \end{minipage}
    \caption{Additional Novel-View synthesis qualitative evalutation on DTU \cite{aanaes2016large} using $3$ source images. We outperform SOTA MVSGaussian \cite{liu2024fast} and exhibit sharper view synthesis results, particularly on object boundaries.}
    \label{fig:nvs_grid_additional}
\end{figure*}

In this section, we present additional qualitative comparisons with MVSGaussian \cite{liu2024fast} in Figure \ref{fig:nvs_grid_additional} to demonstrate the superior performance of our method on the DTU \cite{aanaes2016large} dataset. Our final adapted model exceeds the novel-view capabilities of its initial backbone to achieve SOTA performance and provides good novel-view extrapolation results compared to the generalizable reconstruction networks, which generally exhibit poor NVS performance (as pointed out in GeoTransfer \cite{jena2024geotransfer} supplementary). We notice that qualitatively, in the sparse $3$ input views setting, we are sharper than previous SOTA MVSGaussian \cite{liu2024fast}, with lesser artifacts. This demonstrates the robustness of our method on the task of novel-view synthesis, apart from also displaying \textbf{state-of-the-art} results on surface reconstruction. 

\section{Additional Experimental Details}
\label{sec:exp_details}
In this work, we evaluated two sets of data: DTU \cite{aanaes2016large} and BlendedMVS \cite{yao2020blendedmvs}. For DTU \cite{aanaes2016large}, we follow distinct protocols based on the task's nature, distinguishing between novel view synthesis and surface reconstruction.

\paragraph{\textbf{Metrics}} For the novel view synthesis task involve evaluating PSNR scores, assuming a maximum pixel value of 1 and using the formula $-10 \log_{10}$(MSE). Additionally, we employ the scikit-image's API to calculate the Structural Similarity Index (SSIM) score and the pip package lpips, utilizing a learned VGG model for computing the Learned Perceptual Image Patch Similarity (LPIPS) score. In the context of the surface reconstruction task, we gauge Chamfer Distances by comparing predicted meshes with the ground truth point clouds of DTU scans. The evaluation process follows the methodology employed by SparseNeuS, VolRecon, ReTR, GeoTransfer \cite{long2022sparseneus, ren2023volrecon, liang2024retr, jena2024geotransfer}, employing an evaluation script that refines generated meshes using provided object masks. Subsequently, the script evaluates the chamfer distance between sampled points on the generated meshes and the ground truth point cloud, producing distances in both directions before providing an overall average, typically reported in evaluations. Additionally, two sets of 3 different views are used for each scan, and we average the results between the two resulting meshes from each set of images and report it in the comparison as done in previous methods \cite{long2022sparseneus, ren2023volrecon, liang2024retr, jena2024geotransfer}.

\paragraph{\textbf{DTU Dataset}} The DTU dataset \cite{aanaes2016large} is an extensive multi-view dataset comprising $124$ scans featuring various objects. Each scene is composed of $49$–$64$ views with a resolution of $1600 \times 1200$. We adhere to the procedure outlined in \cite{long2022sparseneus, ren2023volrecon, liang2024retr}, training on the same scenes as employed in these methods and then test on the $15$ designated test scenes the reconstruction task. The test scan IDs surface reconstruction are : $24$, $37$, $40$, $55$, $63$, $192$, $65$, $69$, $83$, $97$, $105$, $106$, $110$, $114$, $118$ and $122$. For surface reconstruction, for each scan, there are two sets of 3 views with the following IDs used as the input views: set-$0$: $23$, $24$ and $33$, then set-$1$: $42$, $43$ and $44$ all scans. We use the training views in the resolution, \ie $640 \times 512$. For the task of novel view synthesis, we use the same split of training and testing views as MVSGaussian \cite{liu2024fast} as well as adopt the same input view set as them for fairness of comparison. 

\paragraph{\textbf{BlendedMVS Dataset}} BlendedMVS \cite{yao2020blendedmvs} is a large-scale dataset for generalized multi-view stereo that consists of a variety of $113$ scenes including architectures, sculptures and small objects with complex backgrounds. For surface-reconstruction, we use $4$ challenging scenes, where each scene has $31$–$143$ images captured at $768 \times 576$.

{
    \small
    \bibliographystyle{ieeenat_fullname}
    \bibliography{main}
}

\end{document}